Article

# Path planning for unmanned naval surface vehicles

**Daniel G. Schwartz**

Department of Computer Science, Florida State University, Tallahassee, FL 32306, USA; schwartz@cs.fsu.edu

**Abstract:** There nowadays is a myriad of approaches to real-time avoidance of fixed obstacles for unmanned surface vehicles (USVs) and, to a lesser extent, also the task of avoiding moving obstacles such as boats, ships, swimmers, and other USVs, but both topics still present challenges. This paper offers novel approaches to both of these problems. It uses a combination of a global path planner, which finds a path from a start point to a goal point that avoids fixed obstacles (given that their locations are known in advance), and a local path planner, which can circumnavigate a moving obstacle (as well as any previously unknown fixed obstacles). The global planner is novel in that it employs a combination of three path planners, one known in the literature as Grassfire, one that is a new modification of Grassfire, and one that is a new, and arguably more intuitive, version of the well-known Probabilistic Roadmap. The local planner is novel in that it employs a higher-level decision logic based on its observations regarding the direction of movement of the obstacle relative to the USVs global path. This logic enables the USV to determine the best strategy for avoiding the obstacle by systematically routing the vehicle behind the obstacle rather than running parallel to it until the opportunity to pass appears. Simulations are provided that validate these claims. For comparison with other systems, the simulations include an implementation of the well-known D* algorithm, and the discussion covers additional dynamic path planning systems, which, like D*, do not necessarily route the vehicle behind the moving obstacle.

## 1. Introduction

There nowadays is a myriad of approaches to robot path planning, including many concerned with unmanned surface vehicles (USVs). This paper offers another approach, one which offers advantages over those that have gone before. Specifically, this work addresses the problem of routing a USV through an environment that is populated with both fixed and moving obstacles (ships, boats, swimmers, other USVs) and, with regard to the latter, routing the vehicle behind the obstacle, rather than in front of or alongside it, whenever this is appropriate, thereby avoiding the possibility of a collision. This invokes a higher-level decision logic that takes into account the position and trajectory of the obstacle relative to that of the USV.

This work specifically addresses surface vehicles traveling on land or water, inasmuch as it considers a two-dimensional operational environment. However, the methods developed here can easily be generalized to three dimensions, so that the same methods can apply also to both aerial and underwater vehicles.

The solution method uses a combination of a global path planner, which finds a path from a start point to a goal point while avoiding known fixed obstacles, and a local path planner, which can detect and circumnavigate a previously unknown moving obstacle. The global planner is novel in that it employs a combination of three path planners known as Grassfire (GF), Modified Grassfire (MGF), and Recursive Probabilistic Roadmap (r-PRM). These work with a cellular grid overlaying the operational environment. Grassfire (also known as Flood Fill, Wave Front, and originally as Cost Wave Propagation) was introduced by Dorst [1]. This finds a least-cost path that follows cell grid lines. An advantage of GF is that it is guaranteed to find a path from the start to the goal, as long as such a path is possible. A drawback is that, in following grid lines, it entails numerous right-angle turns. This motivated augmenting GF with MGF and r-PRM so as to produce a smoother path.

Another contribution of this paper is the r-PRM algorithm, which comprises a recursion-based version of the Probabilistic Roadmap (PRM) due to Kavraki et al. [2]. PRM has enjoyed a wide variety of applications. The newer version being presented here is, in our opinion, intuitively simpler and easier to understand, and it typically is faster. It does have the limitation, however, that the length of the path that can be generated depends on the available size of the computer's runtime stack. This limitation can be overcome, however, by converting to an iteration-based implementation, as discussed in Section 5 (last paragraph).

Both PRM and r-PRM algorithms work with an operational environment containing fixed obstacles and seek to find an obstacle-free path from a given start node to a goal node. They both begin with a random sampling of the obstacle-free space, then use this sampling to create a graphical network reaching from the start to the goal, and then apply Dijkstra's algorithm [3] to find the shortest path through this network. The graphical network is termed the *roadmap*. Both methods require that the sample set be sufficiently large to provide a connected obstacle-free map, and can fail if it is not. The two algorithms differ in how the roadmap is constructed. If the sampling provides a complete map from the start to the goal, both algorithms find a path, and their results are essentially equivalent.

MGF is a simple modification of GF that includes diagonals through grid cells. This can produce a shorter path than GF, but it has the drawback that it can run dangerously close to, actually touching, the fixed obstacles. In our simulation, this has been addressed by adding a one-cell margin to the fixed obstacles for the purposes of both GF and MGF.

Once the GF and MGF paths have been generated, one then takes the polygon formed by these and, within this, applies r-PRM. This produces the desired global path.

The local path planner is triggered when the USV comes into close proximity with a moving obstacle. It is assumed that the algorithm is implemented on-board the USV and works in conjunction with sensors that can detect the position and direction of movement of the moving obstacles. The decision regarding how to avoid the obstacle considers (i) whether the obstacle is on the left or right of the global path, (ii) whether the obstacle is moving toward or away from the global path, and (iii) if the obstacle is moving toward the global path, whether it is moving toward a point behind or in front of the USV. The details of this logic and how it is implemented are provided in the following. We believe that this approach is new, especially in that it takes into account the direction of movement of the obstacle, rather than simply treating it as a newly discovered fixed obstacle.

The latter approach is taken by most other systems that address path finding in dynamic environments. These include the well-known D* algorithm [4], which, for illustrative purposes, has been implemented along with our own system in the following simulations. This also applies to systems based on potential field methods, such as E* [5]. It deserves mentioning that such systems do obey the COLREGS [6] "Rule 15, Crossing Situation (a) When two power-driven vessels are crossing so as to involve risk of collision, the vessel which has the other on her starboard side shall keep out of the way and shall, if the circumstances of the case admit, avoid crossing ahead of the other vessel". Note, however, that this rule leaves open exactly how the USV is supposed to do this.

The present system adopts the policy of always routing the USV behind the moving obstacle whenever the obstacle is detected as moving across the path of the USV. In this respect, however, it deserves mention that the methods devised in [7], which are implemented in the well-known MOOS-IvP simulation framework [8], allow for the USV to cross in front of the obstacle when this is warranted by their relative trajectories and speeds (velocities), i.e., when the obstacle is moving slowly relative to the USV. Moreover, whereas [9] does not discuss global path planning or the task of circumnavigating a moving obstacle, it does generally provide a more fine-grained and detailed analysis of other requirements for a local planner, e.g., how to respond when the USV is overtaking a moving obstacle from the rear. For this reason, it would be appropriate for a future work to incorporate those algorithms into the decision logic provided by the present framework.

This paper combines and expands work previously published as [10] and [11] and which comprises the essential content of Chowdhury's doctoral dissertation [12]. The present author served as Chowdhury's doctoral advisor. A five-minute video of a simulation run and the complete Python code for the simulator can be downloaded at https://github.com/danielgschwartz/USV_Simulation.

The organization of the paper is as follows. Section 2 discusses related research in path planning and autonomous vehicles. Section 3 describes the original PRM algorithm. Section 4 presents the r-PRM algorithm. Section 5 provides experimental results comparing the two methods. Section 6 discusses the global path planner. Section 7 discusses the local path planner. Section 8 provides some experimental results that shows how the proposed local path planner with r-PRM can serve as a complete on-board path planner for real-time obstacle avoidance. This compares the results of our system with that of D*. Section 9 discusses prospects for future research. Section 10 provides some concluding remarks.

## 2. Related work

Some well-known path finding algorithms are A* [13], D* [4], Rapidly-Exploring Random Trees [14], Potential Field Method [15], as well as the two referenced in the foregoing as GF and PRM.

As mentioned, PRM was introduced by Kavraki et al. [16]. Kavraki et al. [2] further showed that PRM is probabilistically complete in the sense that the probability of failure decays to zero exponentially with the number of samples used in the construction of the roadmap. Amato et al. [17] introduced OBPRM: Obstacle-based PRM, by incorporating a visibility graphs sampling method [18] with PRM, which reduces the burden of the high amount of collision checking. Hsu et al. [19] resolved the issue of dynamic threats with PRM by introducing the concept of a "milestone", which creates a real-time graph connecting the start and goal points. Yan et al. [20] implemented the octree-structure PRM which combines PRM with a corridor map sampling method [21]. Khatib [22] introduced the potential field method (PFM) for solving the path planning problems. Such methods have the drawbacks that they are incomplete and prone to drop into local minima. Connolly et al. [23] resolved this local minima problem by incorporating the Laplace equation with PFM. Similarly, Rimon et al. [24] combined PFM with a Morse function having a single minimum to form a stable robot navigation method, thus jumping out of the local minima. Barraquand and Latombe [25] showed that a randomized path planner (RPP) can solve the local minima problem by executing random walks. However, the probability that any random walk finds its way through a narrow passage is almost zero.

LaValle [14] implemented the sampling-based Rapidly Exploring Random Trees (RRT) as a tool for the path planning problem and, subsequently, Yershova et al. [26] proposed an improved version of RRT called DDRRT which overcame the local minima problem of RRT. Similarly, Karaman and Frazzoli [27] proposed an improved version of RRT known as Rapidly Exploring Random Graph (RRG). RRG allows multiple robots to operate simultaneously. Karaman and Frazzoli further improved RRG by pruning bad connections from the RRG graph, resulting in a tree known as RRT-Star (RRT*). RRT* has low time complexity compared to RRT and RRG. Erinc and Carpin [28] combined a genetic algorithm (GA) with the RRT method to achieve a faster convergence speed for the path planning problem. However, RRT cannot do replanning, since it is a single query approach that always returns a single path to the goal instead of many feasible paths. Shamos and Hoey [29] first proposed the Voronoi diagram; subsequently, Luchnikov et al. [30] improved its construction method and applied the improved version of Voronoi diagrams for the 3D path planning problem. Roos and Noltemeier [31] further improved the Voronoi diagram by adding a bound to the Voronoi channels to deal with moving objects, thus making the Voronoi diagram suitable for path planning in a 3D space that has both fixed and moving obstacles. This approach generates a global graph or local graph as with RRG and PRM, but it cannot generate the shortest path by itself. Liu and Zhang [32] proposed an improved version of the Voronoi diagram combined with Dijkstra to find the shortest path for the path planning problems.

Path planning algorithms (both global path planners and local path planners) can be combined for mutual benefit and significantly impact the outcome of a mission planning operation. Dalpe and Thein [33] did an exploratory study to determine which path planning algorithm is a good fit as a global path planner when integrating with a local path planner. Later Dalpe et al. [34] implemented a hybrid global and local path planner using the combination of A-star (A*) [35] and potential field method (PFM) [22] where the former acts as a global path planner and latter acts as local path planner.

Chowdhury and Schwartz have previously explored various combinations of some of these for path planning for unmanned underwater vehicles (UUVs) with special focus on the task of avoiding moving obstacles (boats, ships, etc.) [36, 37]. Since that work has not considered depth information, however, it more correctly applies to surface vehicles or underwater vehicles operating at a constant depth.

In the ocean environment, factors such as weather, waves, currents, international regulations for preventing collisions at sea (COLREGS), and several other challenges must be taken into account. In this regard, the potential field method (PFM) [22,38,39] has been chosen by many researchers because of its elegance and simplicity. Song et al. [49] combined PFM with a multi-layered fast marching (MFM) method to deal with factors like currents and wind. However, sometimes PFM gets stuck in a a local minimum where the USV becomes trapped inside a U-shaped obstacle, or when facing a large wall-like obstacle [41, 42].

The rolling window method [43–45] is commonly used to deal with the local minimum conditions. The main idea is to establish a dynamic window based on the current location of the USV and have the path planner only model the portion of the mission environment that falls into the window area. This approach can reduce the amount of model data necessary to meet real-time requirements. Promising results were found when the local path planner using the rolling window method is combined with other methods such as collision prediction in [9, 46], as well as a fuzzy control method [47]. However, one still has the local minimum problem, especially when the size of the trap/local minimum is larger than the dynamic window size or the range of the USV's sensors. In this case, the USV may wander in the trap and waste energy for an extended period of time.

Thus, a local path planner needs robustness with respect to these challenging factors. Several works have addressed this [48–50]. A key issue for the local planner is to deal with real-time obstacle avoidance. Miotto et al. [51] has proposed an on-board path replanning process for the Manta test vehicle which uses D* algorithm [4] for path replanning to avoid the moving obstacles. That approach can avoid moving obstacles, but it does not necessarily route the USV behind the obstacle.

## 3. The PRM algorithm

Let us begin with an exposition of the original PRM paper [16]. What is presented here is a detailed description of our version of their algorithm, which, for the purposes of a Python implementation, incorporates some minor modifications.

The work begins with a *configuration space*, where each point in this space represents a configuration of the robot. The objective is to construct a path through this space from some *start* configuration to a *goal* configuration. In their examples, the robot is an articulated robotic arm with an arbitrary number of revolute joints. In our application, the configurations are the locations of a USV in some predefined operational environment. The details of this space should be immaterial, however, as the algorithm as presented is generic and should apply equally to both kinds of configurations.

It is assumed that there is a *distance measure* $\delta$ giving the distance between any two configurations. The examples take the operational environment as a two-dimensional space with the usual Cartesian coordinates, and it takes $\delta$ as the Euclidean distance.

The operational environment is divided into a *free space* and an *obstacle space*, where the latter are configurations that are to be avoided. So, more exactly, the objective is to find a path from the start node to the goal through the free space, so as not to collide with any obstacles.

For purposes of the present operational environment, it is natural to refer to the configurations as *points* or *nodes* identified by their $(x, y)$ coordinates.

The first step of the PRM process is to take a random sampling of points in the free space. The application further requires that all such selected points be some specified *min distance* from any obstacles. This is called the *sample set*. These sample points exclude the start node and the goal node. The sample set is configured as an indexed list, $S$. The start node is inserted at the beginning, and the goal node is appended to the end.

The next step is to create the road map. Pseudocode describing the road map construction step, adapted from [2], is shown in Algorithm 1, where "no collision" means that the edge $(c, n)$ does not intersect any obstacle.[1]

**Algorithm 1:** PRM Roadmap Construction Step

**input :** Sample points $S$
**output :** Roadmap $E$

```
1: N ← first sample point
2: E ← ∅
3: for each next point in S
4:     c ← the next point in S
5:     N_c ← a set of candidate neighbors of c
             chosen from N
6:     N ← N ∪ {c}
7:     for all n ∈ N_c, in order of increasing
             δ(c, n) do
8:         If (c, n) ∉ E ∧ no collision then
9:             E ← E ∪ {(c, n)}
```

Thus, the roadmap is defined as a graphical network consisting of a collection of nodes, $N$, and a collection of edges, $E$, connecting the nodes. As indicated in the pseudocode, $N$ is initialized with the first sample point in $S$ (which has been specified to be the start node). Then, for each next point in $S$, find its $k$ nearest neighbors in $N$ (or as many as possible if there are fewer than $k$ points in $N$), and, for each such neighbor, if the edge connecting the point to the neighbor does not intersect an obstacle, this edge is added to $E$. This continues point-by-point through the set $S$ until the goal node is reached. When this happens, $E$ is the *road map*.

Once such an $E$ has been constructed, where $|S|$ is the size of $S$, create an $|S| \times |S|$ reachability matrix and prefill it with the Python null value $\mathrm{None}$. Then record all the road map edges into this matrix by putting the distances between the edge end points into the corresponding matrix cells. Specifically, where $(p_1, p_2)$ is an edge in $E$, and $a$ and $b$ are the indices of $p_1$ and $p_2$ in the list $S$, the matrix cells $(a, b)$ and $(b, a)$ are assigned the value $\delta(p_1, p_2)$. Then apply Dijkstra's algorithm [3] to this matrix. This well-known algorithm takes as input any connected graphical network (here represented as a reachability matrix) and any node in that network, and finds the shortest path from that node to each of the other nodes. The application starts with the designated start node and allows the algorithm to run until it reaches the designated goal node and then returns just this discovered path from the start to the goal. For the simulation result shown in **Figure 1**, the PRM sample set size was 500, including the start and goal nodes, and the number of neighbors $k$ was 16. This algorithm typically produces an acceptable path from the start to the goal, although it might not be optimal (shortest).

## 4. The recursion-based PRM algorithm

As with PRM, this begins with an operational environment that is divided into a *free space* and an *obstacle space*, some designated *start* and *goal* nodes, and a distance measure $\delta$, which again is the Euclidean distance. Exactly as in the foregoing, create a sample set with inserted start and end nodes and configure it as an indexed list $S$. Then, also as before, create an $|S| \times |S|$ matrix, to serve as a reachability matrix, and prefill all the cells with the Python null value $\mathrm{None}$. Then, starting with the start node, proceed to fill in cells of the matrix according to a recursive procedure as follows. Given a *focus node* $p$, the start node being the first focus node, find its $k$ nearest neighbors, $p_1, \ldots, p_k$, in the sample set, and that have not already been visited. If there are not $k$ such points available, take as many as you can find. Ignore those for which the connecting edge intersects an obstacle. Let $a$ be the index of $p$ in $S$. For each neighbor $p'$, let $b$ be the index of $p'$ in $S$, and assign the matrix cells $(a, b)$ and $(b, a)$ the value $\delta(p, p')$. Now recursively repeat this with each $p_k$ as the focus node. Keep doing this until one of the neighbors is the goal.

The reachability matrix at this point describes a network of links connecting the start node to the goal and serves as the *road map*. This is shown in Algorithm 2, where $S$ is the sample point list, $R$ is the reachability matrix, and $\delta$ is the distance measure. Next apply Dijkstra's algorithm as before to find the shortest path through this network from the start to the goal.

As with the PRM simulation, the r-PRM simulation employs a sample set of 500 including the start and goal nodes, and the number of neighbors used in creating the reachability matrix was 16. For each run, the simulation used the same sample set for both PRM and r-PRM so as to compare their results.

Also as with PRM, this process can fail to find a path from the start to the goal if there are not enough sample points to create an obstacle-free path. However, this can be remedied as with PRM by increasing the size of sample set. **Figure 1** shows a typical run of the two algorithms in a somewhat complex operational environment. Here the results are almost identical, but this is often not the case. Depending on the random samplings, PRM and r-PRM can produce very different paths. Nonetheless, as shown in **Table 1**, the path lengths are essentially identical, for which reason the two methods are considered to be equivalent.

**Algorithm 2:** r-PRM roadmap construction step

**input** : Sample points $S$

**output** : Roadmap $R$

1: $R \leftarrow \infty$
2: $p \leftarrow S[0]$
3: $a \leftarrow index\ \ of\ \ p\ \ in\ \ S$
4: **For all** $unvisited\ k\ nearest\ \ neighbors\ \ p'\ of\ p$ **do**
5:     **If** $(p, p')\ \ not\ \ intersect\ \ an\ \ obstacle$ **then**
6:         $b \leftarrow index\ \ of\ \ p'$
7:         $R_{a,b} \leftarrow \delta(p, p')$
8:         $R_{b,a} \leftarrow \delta(p, p')$
9:         **If** $p'\ \ is\ \ goal$ **then**
10:             **break**
11: **For all** $unvisited\ k\ nearest\ \ neighbors\ \ p'\ of\ p$ **do**
12:     $p \leftarrow p'$
13:     $a \leftarrow index\ \ of\ \ p\ \ in\ \ S$
14:     **Go to** $line\ \ 4$

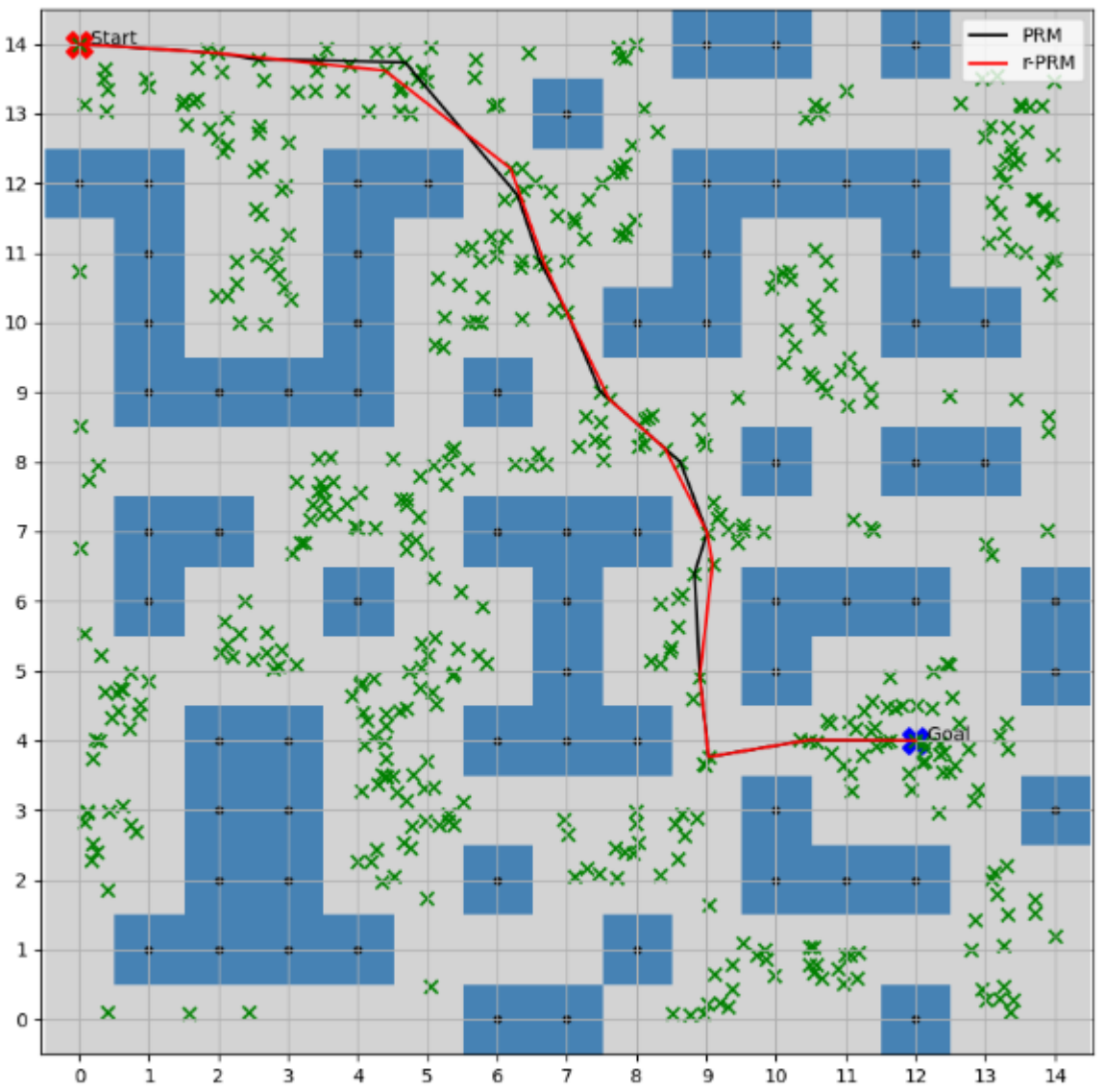

**Figure 1.** PRM versus r-PRM.

**Table 1.** PRM vs r-PRM Path Length in a maze-like operational environment

| No. of Observation[a] | On Maze-like Environment | |
|---|---|---|
| | **PRM Path Length (m)** | **r-PRM Path Length (m)** |
| $n$ = 250, $k$ = 8 | 18.44 | 18.47 |
| $n$ = 500, $k$ = 16 | 18.49 | 18.34 |
| $n$ = 750, $k$ = 32 | 18.23 | 18.20 |
| $n$ = 1000, $k$ = 64 | 18.20 | 18.22 |
| $n$ = 1250, $k$ = 128 | 17.92 | 17.92 |
| $n$ = 1500, $k$ = 256 | 17.89 | 17.89 |
| **Overall** | Mean = 18.20<br>STD = 0.25 | Mean = 18.17<br>STD = 0.23 |

[a] $n$ is the size of the sample set, and $k$ is the number of nearest neighbors joined to each sample point.

**Table 2.** PRM vs r-PRM Roadmap Generation Time in a maze-like operational environment

| No. of Observation[a] | On Maze-like Environment | |
|---|---|---|
| | **PRM Roadmap Generation Time (s)** | **r-PRM Roadmap Generation Time (s)** |
| $n$ = 250, $k$ =8 | 0.40 | 0.48 |
| $n$=500, $k$ =16 | 2.35 | 2.05 |
| $n$=750, $k$ =32 | 11.43 | 6.72 |
| $n$=1000, $k$ =64 | 47.18 | 19.68 |
| $n$=1250, $k$ =128 | 146.97 | 52.75 |
| $n$=1500, $k$ =256 | 354.22 | 153.94 |
| **Overall** | Mean = 93.76<br>STD = 139.06 | Mean = 39.27<br>STD = 59.438 |

[a] $n$ is the size of the sample set, and $k$ is the number of nearest neighbors joined to each sample point.

## 5. Simulation results comparing the two methods

**Figure 1** shows a typical run of the two algorithms in a somewhat complex operational environment. It usually is the case that the algorithms produce essentially the same path. **Figures 2** and **3** show the road maps associated with the respective runs of PRM and r-PRM producing the results in **Figure 1**. The algorithms can produce divergent, but more or less equivalent, paths as shown in **Figures 4** and **5**. **Figure 6** shows a result with a different choice of start and goal nodes. **Figure 7** shows another run with the same start and goal as in **Figure 6**. This has r-PRM producing a somewhat better result than PRM. However, other runs have shown the opposite. **Figure 8** shows an application where the sample set is from a constrained polygon formed by the two path finding algorithms GF and MGF. Note that, in this context, PRM and r-PRM produce almost identical results.

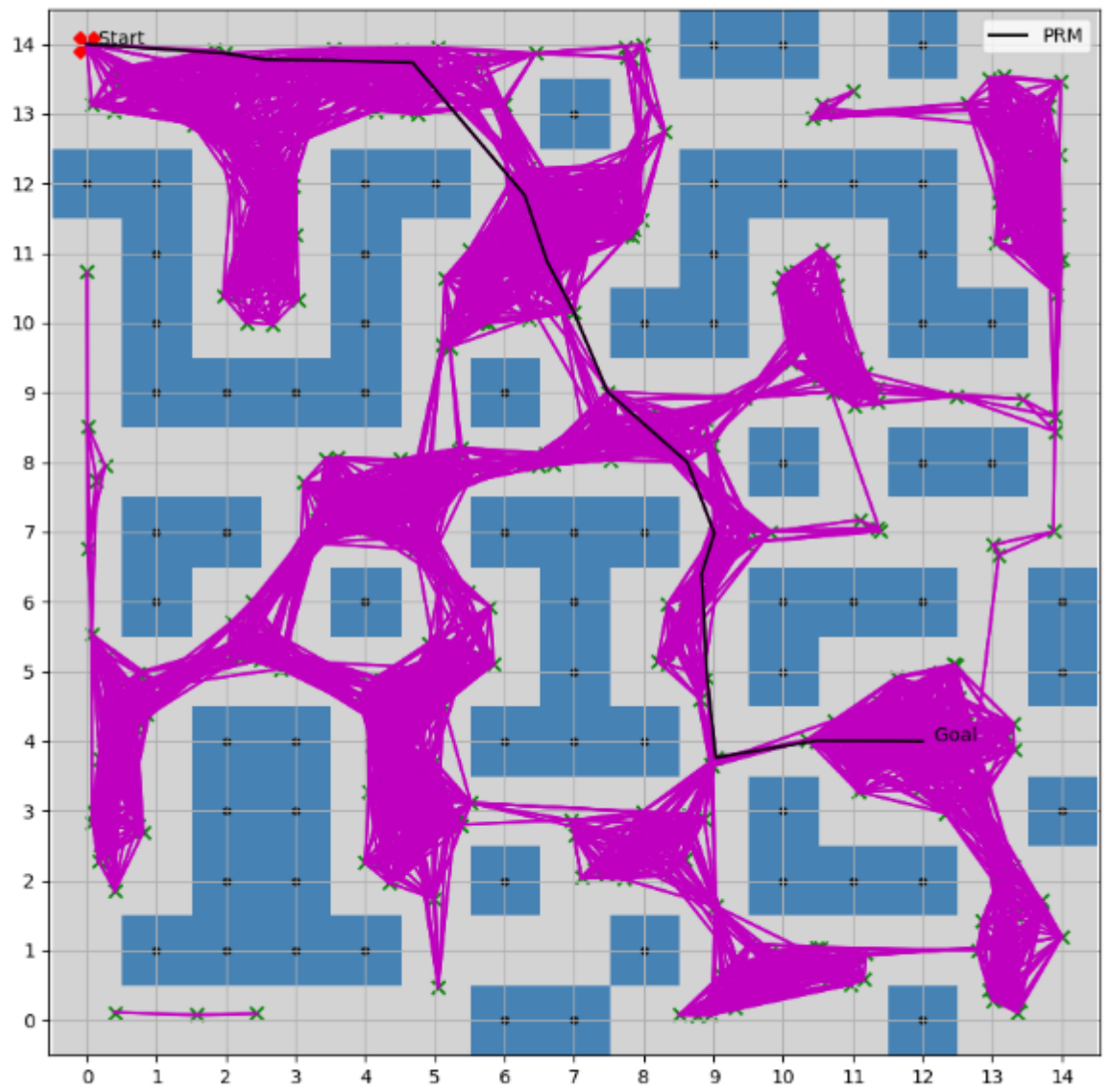

**Figure 2.** PRM roadmap.

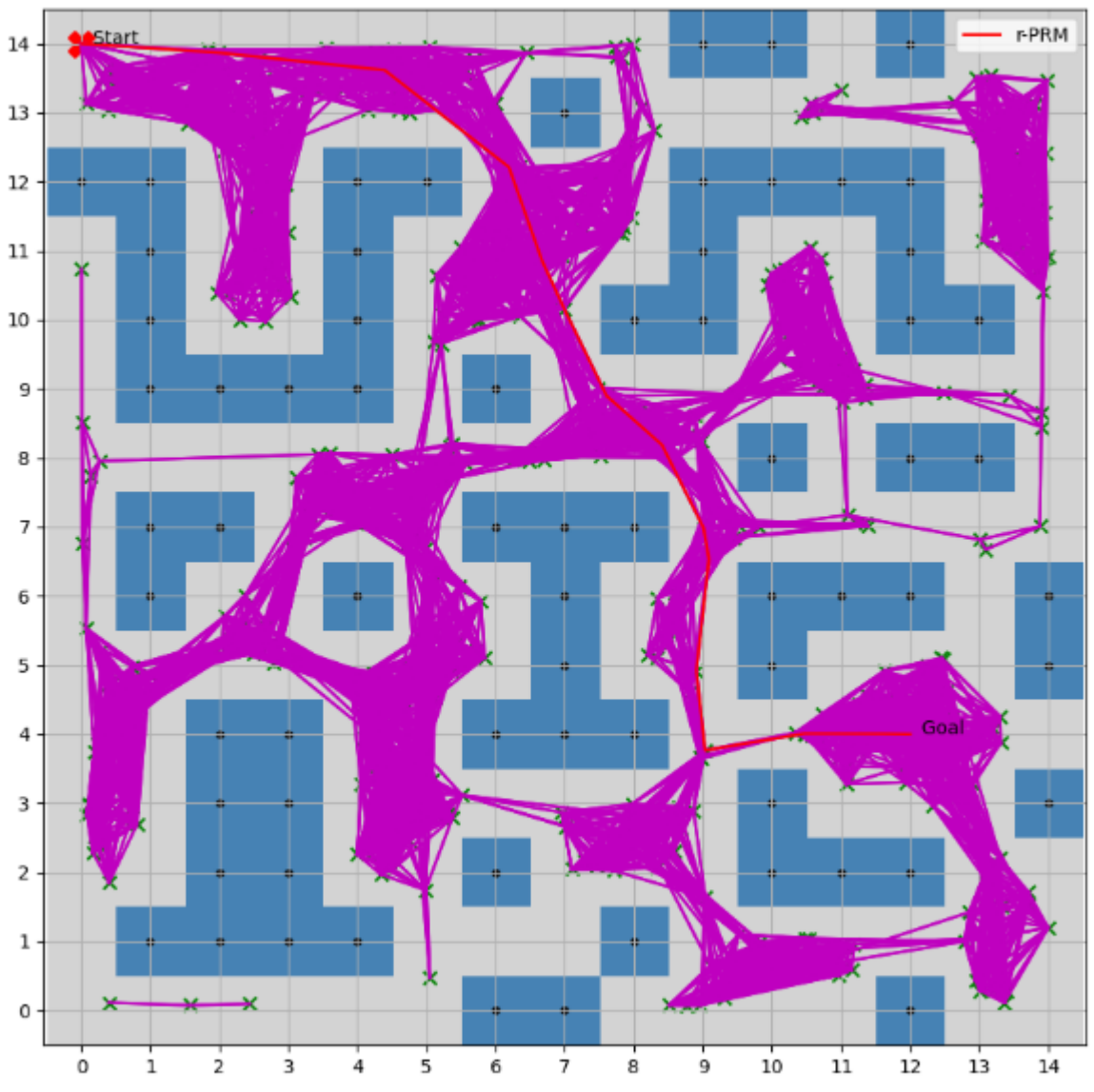

**Figure 3.** r-PRM roadmap.

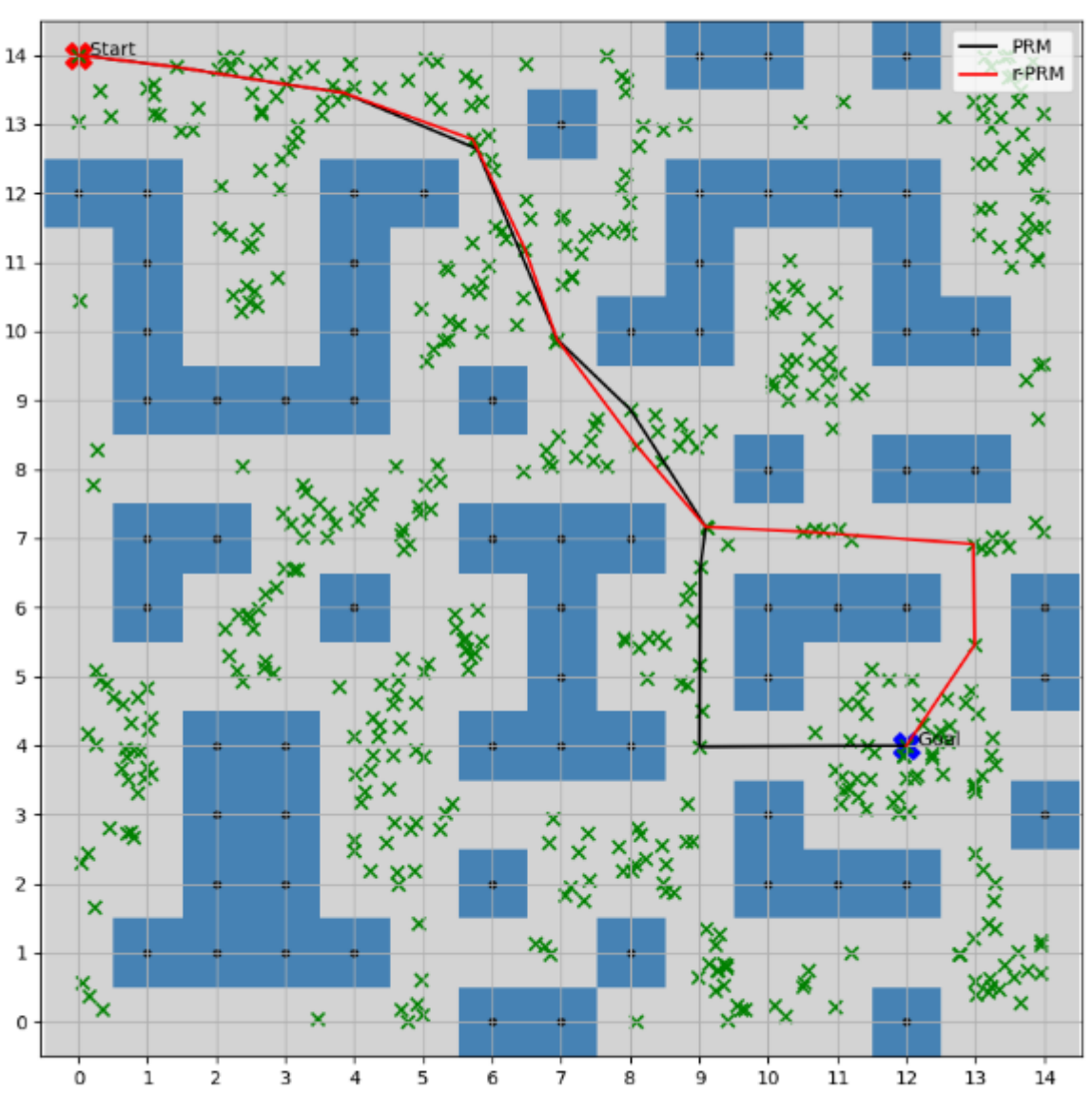

**Figure 4.** PRM versus r-PRM.

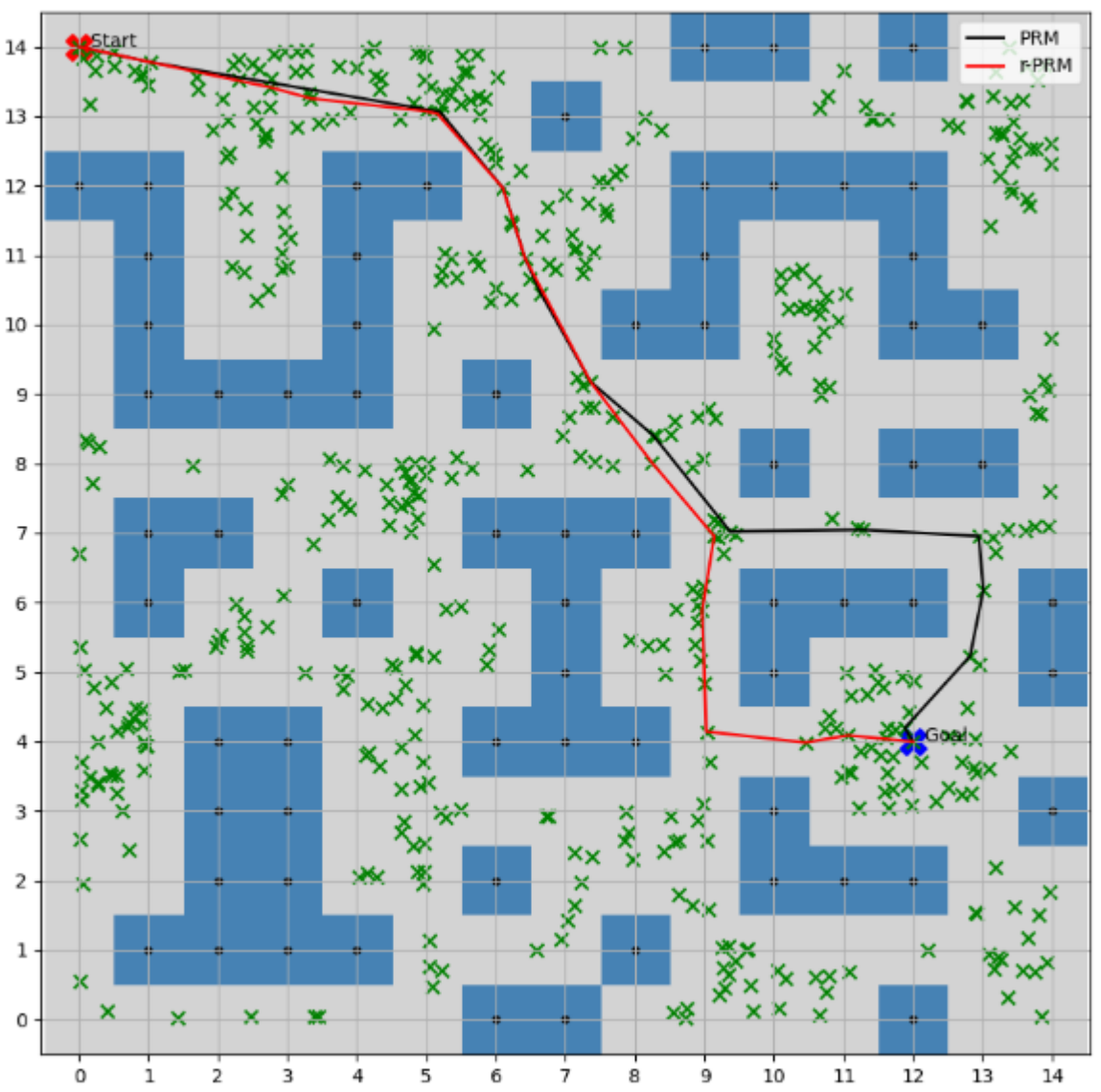

**Figure 5.** PRM versus r-PRM.

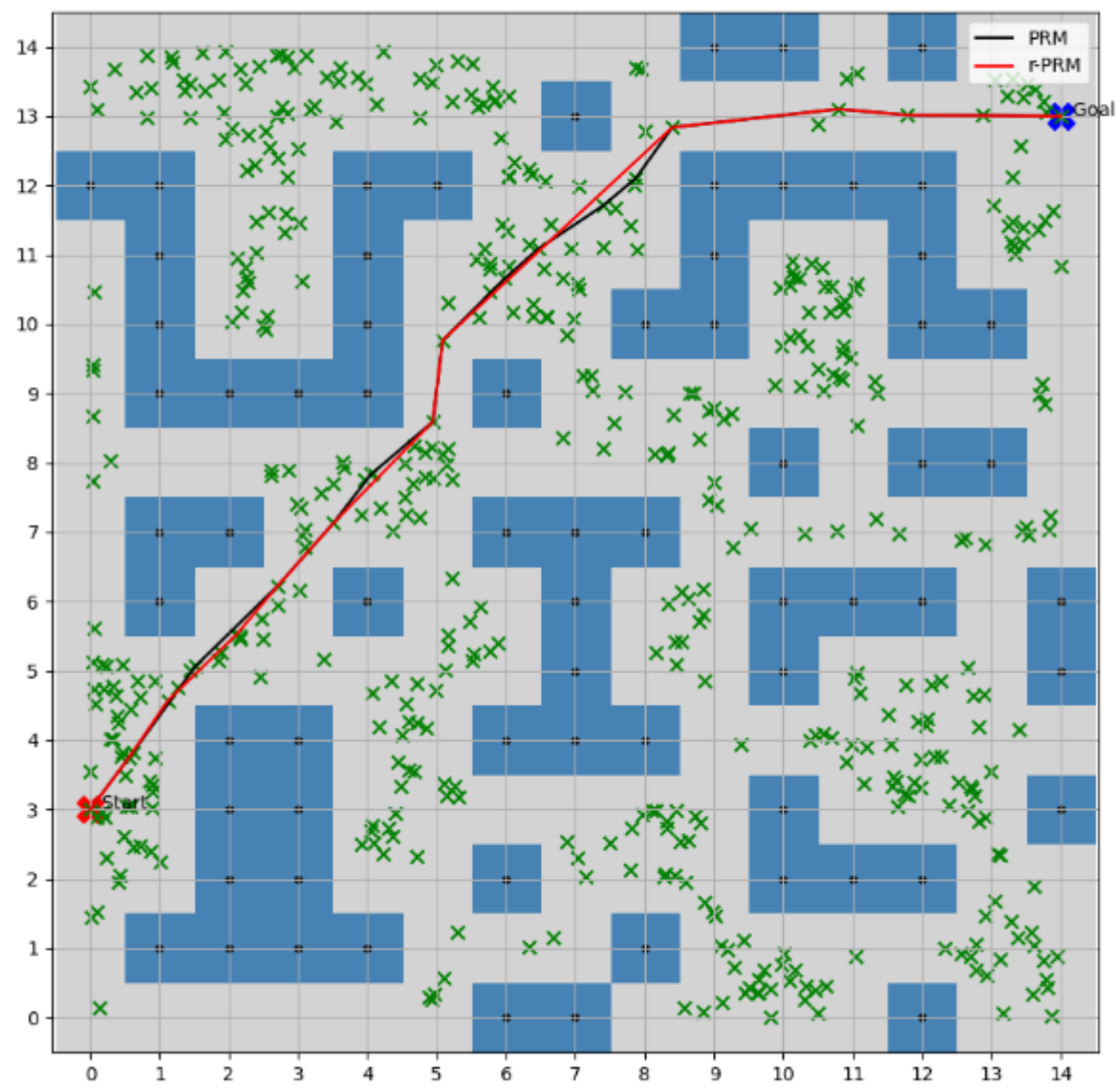

**Figure 6.** PRM versus r-PRM.

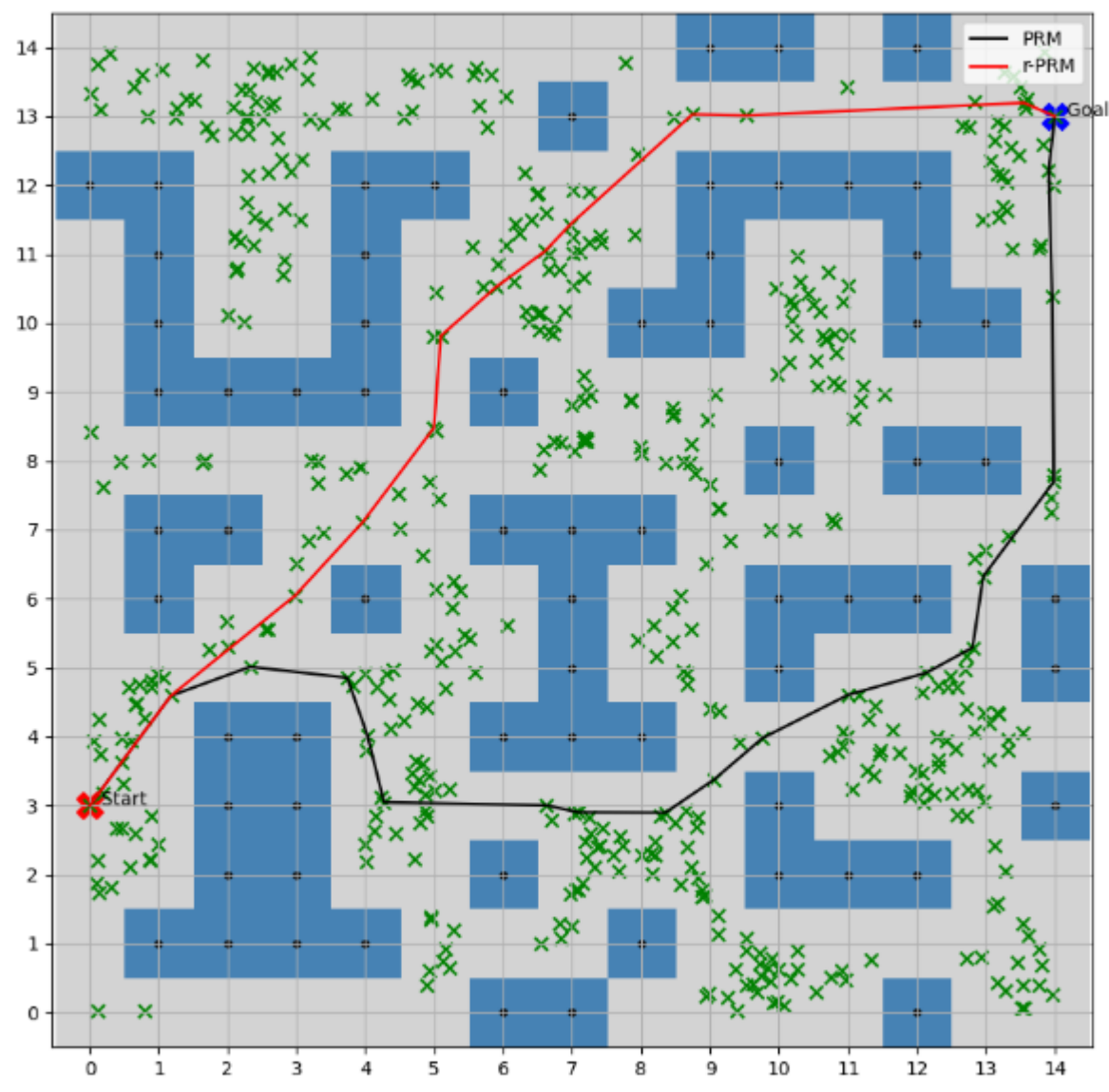

**Figure 7.** PRM versus r-PRM.

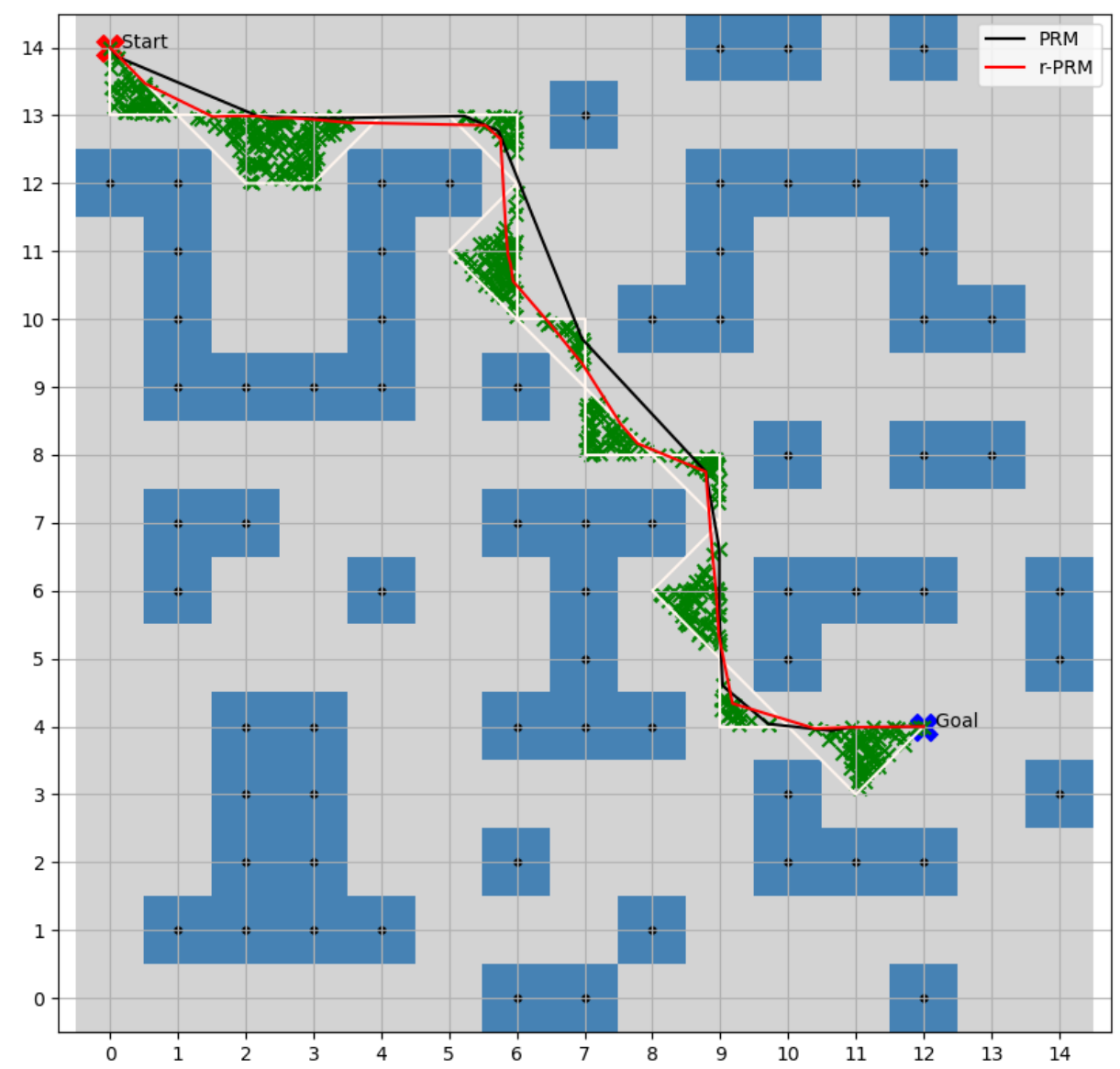

**Figure 8.** PRM versus r-PRM in a constrained polygon.

Thus, overall, the two methods are considered to be essentially equivalent, with the choice of which to employ being a matter of personal preference. Our preference is for r-PRM because it is conceptually simpler and its behavior is easier to visualize. Also, as shown in the following, it usually is faster.

**Table 1** shows the PRM vs r-PRM path length performance in the operational environment of **Figure 1**. **Table 2** shows their respective roadmap generation times in the same environment. **Figure 9** shows an alternative representation of **Table 2**. This explains how the roadmap generation time depends on $n$ and $k$ so that a user can decide how large to make $n$ and $k$ given the available computing resources. The system resource running timeout was set to 1000 s for the RoadMap Generation Time. The results show that PRM failed (i.e., the system crashed due to system resource timeout) but r-PRM did not, when considering relatively a large value such as for $n$=2000 and $k$ =750.

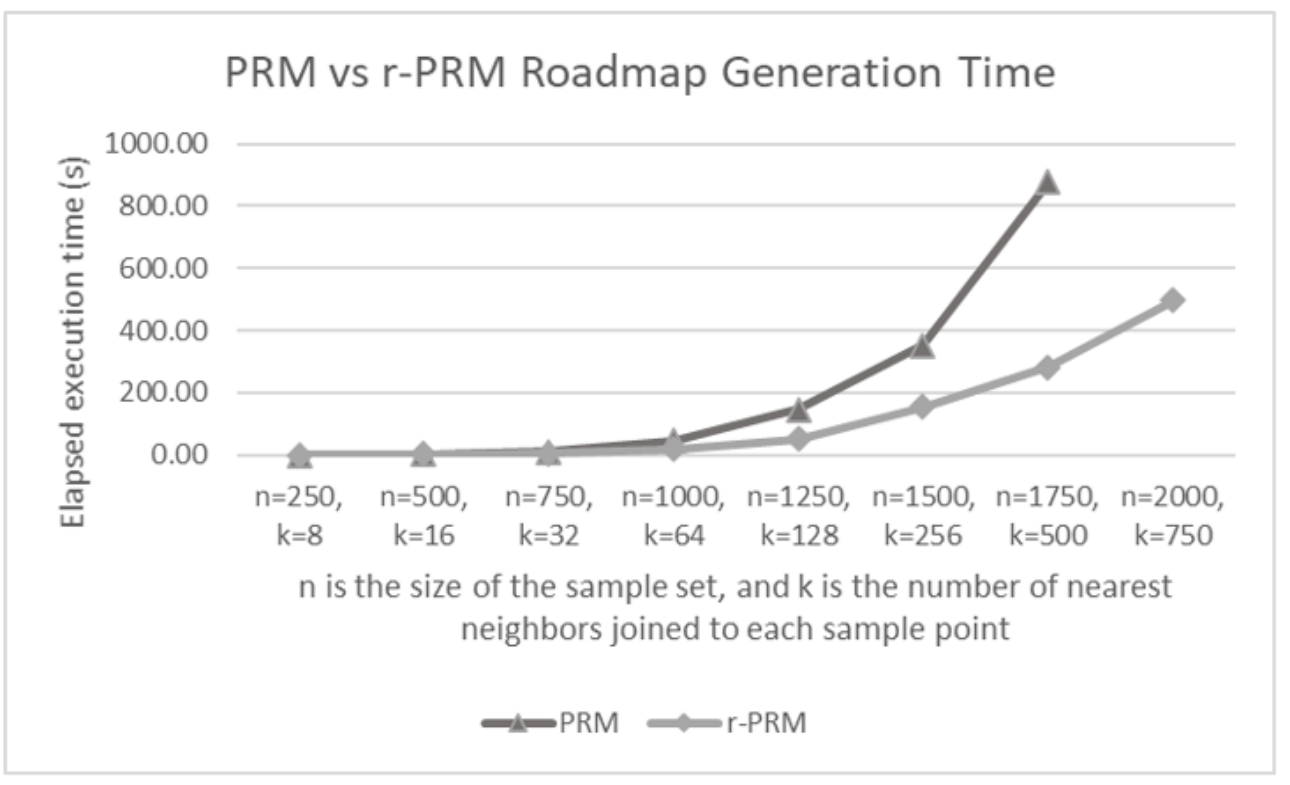

**Figure 9.** PRM versus r-PRM roadmap generation time.

One caveat, however, is that r-PRM is limited by the depth of recursion allowed by the programming language and system on which the algorithm is being implemented. We chose Python because of its packages for creating graphics (the figures). But Python has a recursive depth limit of 1000, which makes it unsuitable for large sample sets.

In languages such as C++ and Java, on the other hand, the recursive depth is limited only by the size of the runtime stack, and this can be adjusted to maximally utilize the available memory.

Having said this, however, it is known that any recursive algorithm can be formulated using only iteration. For example, see [52]. Thus, the same methodology as described here can be implemented in such a way as to overcome the limitation imposed by the depth of recursion. In addition, an iterative version may be expected to run faster as it eliminates the overhead of successive recursive method calls. Thus, this is a further way in which the performance may be improved. The recursive approach was adopted here because it lends itself more intuitively to this particular application.

## 6. Global Path Planner

The global path planning algorithm has three phases: I, Polygon generation; II, Roadmap generation within the polygon (i.e., applying r-PRM); and III, Optimal pathfinding from the roadmap inside the polygon (i.e., applying Dijkstra's algorithm). Pseudocode for the sequential execution of these phases is shown as Algorithm 3.

**Algorithm 3:** Global path finding

```
   input  : worldmap, start, and goal
   output : global path returned: A list of x and y coordinates formed a route from start to goal
 1 path_1 ← GF(worldmap, start, goal) {4-neighbor connectivity};
 2 path_2 ← MGF(worldmap, start, goal) {8-neighbor connectivity};
 3 pn ← Polygon(path_1 + path_2);
 4 while pn.area do
 5     sample_x ← [] {empty_list};
 6     sample_y ← [] {empty_list};
 7     obkdtree ← scipy.spatial.cKDTree(np.vstack((ox, oy)).T){ox, oy are obstacle xy-coordinates};
 8     while len(sample_x) < MAX_SAMPLES do
 9         point_in_poly ← get_random_point_in_polygon(pn);
10         px, py ← np.array(point_in_poly);
11         dist,  ind ← obkdtree.query((np.array([px, py])
12             .reshape(2, 1)).T,  k=1);
13         if dist[0] >= robot_size then
14             sample_x.append(px);
15             sample_y.append(py);
16     fskdtree ← scipy.spatial.cKDTree(np.vstack((sample_x,
17         sample_y)).T);
18     rprm_road_map ← create_rprm_road_map_matrix((start[0],
19         start[1]), sample_x, sample_y);
20     if path ← dijkstra_search(start, goal, rprm_road_map) then
21         break;
22     return path {global path returned};
```

### 6.1. Phase I: Polygon generation

Line 1 of Algorithm 3 calls the GF algorithm, the details of which are shown in Algorithm 4 and are discussed in the following. Line 2 calls the MGF algorithm, also discussed in the following. Line 3 combines the GF and MGF results to form the desired polygon.

**Algorithm 4:** $Grassfire_Algorithm$ with 4-neighbor connectivity

```
    input  : worldmap, start, and goal
    output : path returned: A list of x and y coordinates formed a route from start to goal
 1  closedlist ← set(); closed_list ← {}; oheap ← [];;
 2  neighbors ← [(0,1), (0,−1), (1,0), (−1,0)] {4-neighbor connectivity};
 3  distance ← {goal : 0};
 4  heapq.heappush(oheap, (distance[goal], goal))
 5  while oheap do
 6      current ← heapq.heappop(oheap)[1] ;
 7      if current == start then
 8          path ← [];
 9          while current in closed_list do
10              path.append(current);
11              current ← closed_list[current];
12          return path;
13      closedlist.add(current);
14      forall i, j in neighbors do
15          neighbor ← current[0] + i, current[1] + j;
16          temp_distance ← distance[current] + 1;
17          if 0 <= neighbor[0] < worldmap.shape[0] then
18              if 0 <= neighbor[1] < worldmap.shape[1] then
19                  if worldmap[neighbor[0]][neighbor[1]]
                     >= 1 then
20                      continue{Neighbor: Inside the worldmap and raised collision with Obstacle or Buffer};
21              else
22                  continue{Neighbor: Beyond the worldmap's Y-bound};
23          else
24              continue{Neighbor: Outside the worldmap's XY-bound};
25          if neighbor in closedlist then
26              continue;
27          if neighbor not in [i[1] for i in oheap] then
28              closed_list[neighbor] ← current;
29              distance[neighbor] ← temp_distance;
30              heapq.heappush(oheap,
31                  (distance[neighbor], neighbor));
```

Both GF and MGF work with a cellular grid overlaying the operational environment. Our simulations, shown in the following figures, used a $60 \times 60$ grid. The cells are divided into an *obstacle space*, colored blue, and a *free space*, colored grey. Path finding works with two designated grid points that will serve as start and goal nodes. GF begins at the goal node, assigning it a *distance* value of 0. Then it assigns the value 1 to each of the goal's four free-space neighbor points, i.e., the ones connected to the goal by grid cell edges, but excluding points that belong to obstacles. Then, the free-space neighbors of each of these new points that have not yet been assigned a value are assigned the value 2. And so on, at each step assigning the new nodes the value of the current node $+1$, until this process reaches the start node. Then, starting with the start node, the GF path is formed by proceeding backwards toward the goal by selecting nodes with progressively smaller distance values. Note that such a path will always be found, as long as it is possible to reach from the goal to the start via the free space, and this will be a shortest possible path that follows grid cell edges.

MGF is essentially identical to GF, except that it follows diagonals through cells as well as cell edges. Thus, GF and MGF are distinguished from each other simply in that GF uses 4-neighbor connectivity, while MGF uses 8-neighbor connectivity. MGF has the advantage that it produces a path that is shorter than GF, but it has the disadvantage that it can run dangerously close to, actually touching some of the obstacles. Therefore, to prevent the latter, a 1-cell margin, colored grey, has been added to all the shown obstacles for purposes of these two algorithms. This has the effect of creating a 1-cell buffer around all the designated obstacles. The world map that is input to GF and MGF is a $60 \times 60$ array having 0 to designate free-space grid points, 1 to designate obstacle-space grid points, and 2 to designate buffer points.

The pseudocode shown in Algorithm 3 should be self-explanatory to anyone that is familiar with Python.

### 6.2. Phase II: Roadmap generation within the polygon

In this phase, a roadmap is created within the polygon created in Phase I. The polygon area is first sampled in lines 4 through 15 of Algorithm 3 . Here *ox* and *oy* are the lists of $x$ and $y$ grid coordinates of the points in the obstacle space. These are used to construct a k-d tree called *obkdtree*. The k-d tree is a well-known data structure due to [4] that facilitates nearest-neighbor searches. *MAX_SAMPLES* is a predetermined number of sample points to be selected. Random points within the designated polygon are chosen, and each is checked to make sure it is at least a predetermined *robot_size* distance from any obstacle. Our simulation selected 500 such points. The start node and goal node are added to this set. *sample_x* and *sample_y* are indexed lists of the $x$ and $y$ coordinates of these points. These are then encoded in a k-d tree called *fskdtree* in line 16.

The roadmap is generated in line 18 by calling Algorithm 5 . This incorporates the functionality of Algorithm 3 , discussed in Section IV. The pseudocode shown in Algorithm 5 gives the main steps. The output is to be a reachability matrix of size *MAX_SAMPLES* $\times$ *MAX_SAMPLES*. This matrix is initially null filled. As discussed in Section IV, r-PRM starts with some given start node as the current node, say having index $i$ in the sample point lists, finds that node's $k$ nearest neighbors in the sample space, and for each neighbor checks whether the line segment connecting the current node to the neighbor intersects an obstacle, and if not, where the neighbor has index $j$ in the sample point lists, sets the contents of the $i, j$ cell in the reachability matrix to be the length of this line segment. The current node is then removed from the list of potential neighbors, and the process repeats on all the neighbors. This continues recursively until the goal node is reached.

**Algorithm 5:** $create_rprm_road_map_matrix$

```
input  : current_point, sample_x, sample_y
output : roadmap, E
1  current_sample_point_index ← index_of_sample_point(current_point);
2  sample_x.remove(
3    sample_x[current_sample_point_index]);
4  sample_y.remove(
5    sample_y[current_sample_point_index]);
6  fskdtree ← scipy.spatial.cKDTree(np.vstack(
7    (sample_x, sample_y)).T);
8  distance, index ← fskdtree.query(current_point, k = nearest_neighbor_count);
9  row ← index_of_sample_point(current_point);
10 x_1 ← current_point[0];
11 y_1 ← current_point[1];
12 forall i1 in range(index.size) do
13     x_2 ← sample_x[index[i1]];
14     y_2 ← sample_y[index[i1]];
15     column ← index_of_sample_point((x_2, y_2));
16     if not collision(x_1, y_1, x_2, y_2, fixed_obstacle_kdtree) then
17         rprm_road_map_matrix[row][column] ← distance[i1];
18         rprm_road_map_matrix[column][row] ← distance[i1];
19     create_rprm_road_map_matrix((x_2, y_2),
20       sample_x, sample_y);
```

### 6.3. Phase III: Optimal pathfinding from the roadmap inside the polygon

Once the *rpm_road_map_matrix* is completed, it is sent to the *Dijkstra_search* procedure in line 20 of Algorithm 3 . This applies Dijkstra's algorithm [3] to find the shortest path from the start to the goal through the graphical network represented by the reachability matrix. This is the global path.

## 7. Local path planner

This section introduces details of the various algorithms and decision methods that will allow the USV to trigger local path planning at the most opportune time and to route the USV behind the obstacle. Pseudocode for the simulation program is given in Algorithms 6 through 12. Algorithm 6 initiates a run by generating appropriate representations for the global path and the moving obstacle paths. This then invokes Algorithm 7, which moves the USV along the global path, moves the obstacles along their paths, and continuously checks whether the USV is in close proximity with a moving obstacle. The test for nearby moving obstacles is in line 7 using the predefined *min_fixed_dist*. Testing "true" invokes the local path planner in Algorithm 8. This presents the decision procedure for avoiding the moving obstacle. This considers whether the obstacle is on the left or right of the global path, whether the obstacle is moving toward, or away from, or parallel to the global path, and, if moving toward the global path, whether it is moving toward a point behind or in front of the USV. Depending on the combination of these factors, the USV chooses to go either to the left or the right of the USV, as well as whether this will be a "one-step" or a "two-step" movement, after which it then generates a new global path from the last step point to the goal. Algorithm 9 shows how to move one step right. Algorithm 10 shows how to move two steps right. Algorithm 11 shows how to move one step left. Algorithm 12 shows how to move two steps left. This stepping procedure is carried out as follows.

As shown in **Figure 10**, a line is constructed from the USV through the obstacle; another line is constructed perpendicular to this line through the obstacle; and a circle with a predetermined radius is drawn around the obstacle. Then, for example, if the obstacle is on the left of the global path and is moving toward a point in front of the USV, a two-step circumscription is performed by constructing a line segment from the USV to the leftmost point where the perpendicular intersects the circle, another line segment from that point to the farthest point where the line from the USV through the obstacle intersects the circle, and then a new USV path is formed by joining these two line segments together and finally adding a newly computed global path from the second point to the goal. A one-step circumscription only uses the first of the intersection points.

**Algorithm 6:** *initiate_simulation*

**input** : *start*, *goal*, *min_fixed_dist*, *time_step*
**output** : USV starts following rprm_mgf_path

1 Generate a rPRM-MGF path from the start node to the goal node;
2 *rprm_mgf_path* ← representation of the rPRM-MGF path as a sequence of N steps;
3 $ob_i$*_path* ← representation of obstacle $ob_i$ path as a sequence of N steps, for each $i$;
4 *follow_global_path_and_move_obstacles*(*rprm_mgf*
5 *_path*,*start*, *goal*, *min_fixed_dist*, *time_step*);

**Algorithm 7:** $follow_global_path_and_move_obstacles$

**input** : *rprm_mgf_path*, *current_location_of_USV*, *goal*, *min_fixed_dist*, *time_step*
**output** : USV keeps following rprm_mgf_path; otherwise, calls USV_local_path_planner; moving obstacles follow their paths

1 $ux, uy$ ← *current_location_of_USV*;
2 **while** $(ux, uy)$ *!= goal* **do**
3   **for** *each simulated moving obstacle* $ob_i$ **do**
4     $\{i = 1 \ldots, n$ where $n$ is the number of moving obstacles$\}$;
5     *USV_to_ob_dist* ← *distance*$(ux, uy, ox, oy)$;
6     {distance between USV and $ob_i$, where $(ox, oy)$ is the current location of $ob_i$};
7     **if** *USV_to_ob_dist* <= *min_fixed_dist* **then**
8       *USV_localpath_planner*(*rprm_mgf_path*,
9         *time_step , ux, uy, ox, oy, USV_to_ob_dist*, i);
10     **else**
11       $(ux, uy)$ ← start of the next step of the *rprm_mgf_path*;
12       $(ox, oy)$ ← start of the next step of the $ob_i$ path;
13   wait for one *time_step* (both USV and obstacles);

**Algorithm 8:** *USV_local_path_planner*

**input :** $USV_x, USV_y, USV_x_previous,$
$USV_y_previous, ob_x, ob_y, ob_x_previous,$
$ob_y_previous, USV_path$

**output :** $USV_path, tx, ty$

1: $obVersusUSV \leftarrow sign((pux, puy), (ux, uy), (ox, oy))$;
2: $obIsLeftOfUSVPath \leftarrow (obVersusUSV > 0)$;
3: $obIsRightOfUSVPath \leftarrow (obVersusUSV < 0)$;
4: $obIsOnUSVPath \leftarrow (obVersusUSV == 0)$;
5: **if** $slopeUSVPath == slopeObPath$ **then**
6: print("USV path and ob path are parallel");
7: **if** $obIsLeftOfUSVPath$ **or** $obIsOnUSVPath$ **then**
8: print("ob is left of or on USV path");
9: print("go right one step");
10: $USV_path, tx, ty \leftarrow go_right_one_step(USV_x, USV_y, pux, puy, ux,$
$uy, ox, oy, USV_path)$;
11: **return** $USV_path, tx, ty$;
12: **end if**
13: **if** $obIsRightOfUSVPath$ **then**
14: print("ob is right of USV path");
15: print("go left one step");
16: $USV_path, tx, ty = go_left_one_step(USV_x, USV_y, pux, puy, ux,$
$uy, ox, oy, USV_path)$;
17: **return** $USV_path, tx, ty$;
18: **end if**
19: **else**
20: print("USV path and ob path not parallel");
21: **end if**
22: $intersectionPointVersusLeftPerpendicular \leftarrow sign((ux, uy), (rx, ry), (ix, iy))$ {intersection point $(ix, iy)$ of USV path and ob path. reference point $(rx, ry)$ that is to the left of the USV path and on a line perpendicular to the path through the point $(ux, uy)$};
23: $intersectionPointIsBehindUSV \leftarrow (intersectionPointVersusLeftPerpendicular > 0)$;
24: $intersectionPointIsAheadOfUSV \leftarrow (intersectionPointVersusLeftPerpendicular < 0)$;
25: $intersectionPointIsAtUSV \leftarrow (intersectionPointVersusLeftPerpendicular == 0)$;
26: $distancePobFromUSVPath \leftarrow$
$abs((ux - pux) * (puy - poy) - (pux - pox) * (uy - puy))/distanceBetweenPoints(pux, puy, ux, uy)$;
27: $distanceObFromUSVPath \leftarrow$
$abs((ux - pux) * (puy - oy) - (pux - ox) * (uy - puy))/distanceBetweenPoints(pux, puy, ux, uy)$;
28: $obMovingTowardUSVPath \leftarrow (distancePobFromUSVPath > distanceObFromUSVPath)$;
29: $obMovingAwayFromUSVPath \leftarrow (distancePobFromUSVPath < distanceObFromUSVPath)$;
30: $distanceUSVFromObPath \leftarrow$
$abs((ox - pox) * (poy - uy) - (pox - ux) * (oy - poy))/distanceBetweenPoints(pox, poy, ox, oy)$;
31: $USVMovingTowardObPath \leftarrow (distancePUSVFromObPath > distanceUSVFromObPath)$;
32: **if** $intersectionPointIsBehindUSV$ **or**
$intersectionPointIsAtUSV$ **then**
33: **if** $(obIsLeftOfUSVPath$ **or** $obIsOnUSVPath)$ **and**
$obMovingTowardUSVPath$ **and**
$USVMovingTowardObPath$ **then**
34: $USV_path, tx, ty \leftarrow go_right_one_step(USV_x, USV_y, pux, puy, ux, uy,$
$ox, oy, USV_path)$
35: **return** $USV_path, tx, ty$;
36: **end if**

**Algorithm 8:** (*Continued*)

37: **if** $(obIsRightOfUSVPath$ **and** $obMovingTowardUSVPath$ **and** $USVMovingTowardObPath)$ **then**
38: $USV_path, tx, ty \leftarrow go_left_one_step(USV_x, USV_y, pux, puy, ux, uy, ox, oy, USV_path)$
39: **return** $USV_path, tx, ty$;
40: **end if**
41: **return** $USV_path, ux, uy$;
42: **end if**
43: **if** $intersectionPointIsAheadOfUSV$ **then**
44: **if** $obIsLeftOfUSVPath$ **or** $obIsOnUSVPath$ **then**
45: **if** $obMovingTowardUSVPath$ **then**
46: $USV_path, tx, ty \leftarrow go_left_two_steps(USV_x, USV_y, pux, puy, ux, uy, ox, oy, USV_path)$;
47: **return** $USV_path, tx, ty$;
48: **else if** $obMovingAwayFromUSVPath$ **then**
49: $USV_path, tx, ty \leftarrow go_right_one_step(USV_x, USV_y, pux, puy, ux, uy, ox, oy, USV_path)$;
50: **return** $USV_path, tx, ty$;
51: **end if**
52: **end if**
53: **if** $obIsRightOfUSVPath$ **then**
54: **if** $obMovingTowardUSVPath$ **then**
55: $USV_path, tx, ty \leftarrow go_right_two_steps(USV_x, USV_y, pux, puy, ux, uy, ox, oy, USV_path)$;
56: **return** $USV_path, tx, ty$;
57: **else if** $obMovingAwayFromUSVPath$ **then**
58: $USV_path, tx, ty \leftarrow go_left_one_step(USV_x, USV_y, pux, puy, ux, uy, ox, oy, USV_path)$;
59: **return** $USV_path, tx, ty$;
60: **end if**
61: **end if**
62: **end if**

**Algorithm 9:** $go_right_one_step$

**input** : $USV_x, USV_y, pux, puy, ux, uy, ox, oy, USV_path$

**output** : $USV_path, target_x, target_y$

1 $slopeOfUSVObLink \leftarrow (oy - uy)/(ox - ux)$;
2 $slopeOfPerpendicular \leftarrow -1/slopeOfUSVObLink$;
3 $offset \leftarrow offsetFactor * min_fixed_distance$;
4 {intersection points of perpendicular with circle around ob};
5 $intersection_x_1 \leftarrow ox + offset/np.sqrt(1 + slopeOfPerpendicular **2)$;
6 $intersection_y_1 \leftarrow slopeOfPerpendicular * (intersection_x_1 - ox) + oy$;
7 $intersection_x_2 \leftarrow ox - offset/np.sqrt(1 + slopeOfPerpendicular **2)$;
8 $intersection_y_2 \leftarrow slopeOfPerpendicular * (intersection_x_2 - ox) + oy$;
9 {choose intersection point to right of line from USV to ob};
10 **if** $sign((ux, uy), (ox, oy), (intersection_x_1, intersection_y_1)) < 0$ **then**
11 $\quad target_x \leftarrow intersection_x_1$;
12 $\quad target_y \leftarrow intersection_y_1$;
13 **else**
14 $\quad target_x \leftarrow intersection_x_2$;
15 $\quad target_y \leftarrow intersection_y_2$;
16 $new_target_x, new_target_y \leftarrow nearest_grid_point(target_x, target_y)$;
17 $connecting_line_segment \leftarrow get_line_segment_waypoints(ux, uy, new_target_x, new_target_y)[: -1]$;
18 $new_global_path \leftarrow create_new_global_path((new_target_x, new_target_y), goalNode)$;
19 $USV_index \leftarrow USV_path.index((USV_x, USV_y))$;
20 **del** $USV_path[USV_index : len(USV_path)]$;
21 $USV_path \leftarrow USV_path + connecting_line_segment + new_global_path$;
22 **return** $USV_path, new_target_x, new_target_y$;

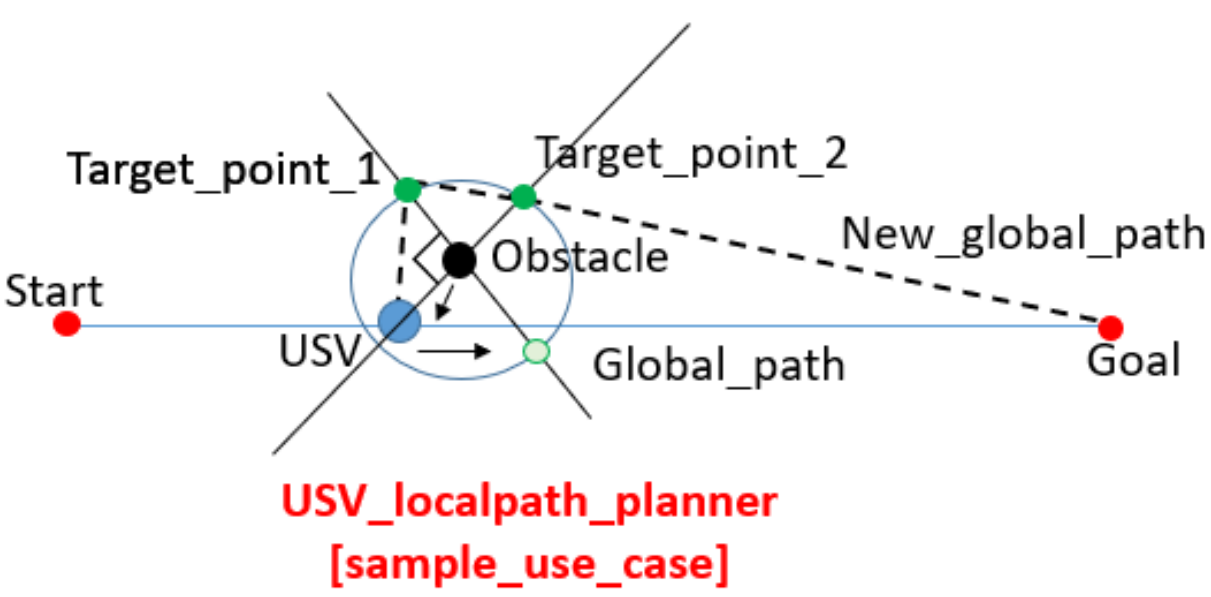

**Figure 10.** Obstacle Avoidance Using A Novel Local Path Planner.

**Algorithm 10:** $go_right_two_steps$

**input :** $USV_x, USV_y, pux, puy, ux, uy, ox, oy,$
$USV_path$

**output :** $USV_path, target_x, target_y$

1: $slopeOfUSVObLink \leftarrow (oy - uy)/(ox - ux)$;
2: $slopeOfPerpendicular \leftarrow -1/slopeOfUSVObLink$;
3: $offset \leftarrow offsetFactor * min_fixed_distance$;
4: {intersection points of perpendicular with circle around ob};
5: $intersection_x_1 \leftarrow ox + offset/np.sqrt(1 + slopeOfPerpendicular * *2)$;
6: $intersection_y_1 \leftarrow slopeOfPerpendicular * (intersection_x_1 - ox) + oy$;
7: $intersection_x_2 \leftarrow ox - offset/np.sqrt(1 + slopeOfPerpendicular * *2)$;
8: $intersection_y_2 \leftarrow slopeOfPerpendicular * (intersection_x_2 - ox) + oy$;
9: {choose intersection point to right of line from USV to ob};
10: **if** $sign((ux, uy), (ox, oy), (intersection_x_1,$
$intersection_y_1)) < 0$ **then**
11: $target_1_x \leftarrow intersection_x_1$;
12: $target_1_y \leftarrow intersection_y_1$;
13: **else**
14: $target_1_x \leftarrow intersection_x_2$;
15: $target_1_y \leftarrow intersection_y_2$;
16: **end if**
17: $line_segment_1 \leftarrow get_line_segment_waypoints(ux, uy, target_1_x,$
$target_1_y)[: -1]$
18: {intersection points of USV-ob line with circle around ob};
19: $intersection_x_3 \leftarrow ox + offset/np.sqrt(1 + slopeOfUSVObLink * *2)$;
20: $intersection_y_3 \leftarrow slopeOfUSVObLink * (intersection_x_3 - ox) + oy$;
21: $intersection_x_4 \leftarrow ox - offset/np.sqrt(1 + slopeOfUSVObLink * *2)$;
22: $intersection_y_4 \leftarrow slopeOfUSVObLink * (intersection_x_4 - ox) + oy$;
23: {choose intersection point to right of line from target_1 to ob};
24: **if** $sign((target_1_x, target_1_y), (ox, oy),$
$(intersection_x_3, intersection_y_3)) < 0$ **then**
25: $target_2_x \leftarrow intersection_x_3$;
26: $target_2_y \leftarrow intersection_y_3$;
27: **else**
28: $target_2_x \leftarrow intersection_x_4$;
29: $target_2_y \leftarrow intersection_y_4$;
30: **end if**
31: $new_target_2_x, new_target_2_y \leftarrow nearest_grid_point(target_2_x, target_2_y)$;
32: $line_segment_2 \leftarrow get_line_segment_waypoints(target_1_x, target_1_y,$
$new_target_2_x, new_target_2_y)[: -1]$
33: $new_global_path \leftarrow create_new_global_path((new_target_2_x,$
$new_target_2_y), goalNode)$;
34: $USV_index \leftarrow USV_path.index((USV_x, USV_y))$;
35: **del** $USV_path[USV_index : len(USV_path)]$;
36: $USV_path \leftarrow USV_path + line_segment_1 + line_segment_2 + new_global_path$;
37: **return** $new_target_2_x, new_target_2_y$;

**Algorithm 11:** $go_left_one_step$

```
input  : USV_x, USV_y, pux, puy, ux, uy, ox, oy,
         USV_path
output : USV_path, target_x, target_y
1  slopeOfUSVObLink ← (oy − uy)/(ox − ux);
2  slopeOfPerpendicular ← −1/slopeOfUSVObLink;
3  offset ← offsetFactor ∗ min_fixed_distance;
4  {intersection points of perpendicular with circle around ob};
5  intersection_x_1 ← ox + offset/np.sqrt(1 + slopeOfPerpendicular ∗ ∗2);
6  intersection_y_1 ← slopeOfPerpendicular ∗ (intersection_x_1 − ox) + oy;
7  intersection_x_2 ← ox − offset/np.sqrt(1 + slopeOfPerpendicular ∗ ∗2);
8  intersection_y_2 ← slopeOfPerpendicular ∗ (intersection_x_2 − ox) + oy;
9  {choose intersection point to left of line from USV to ob};
10 if sign((ux, uy), (ox, oy), (intersection_x_1,
     intersection_y_1)) > 0 then
11     target_x ← intersection_x_1;
12     target_y ← intersection_y_1;
13 else
14     target_x ← intersection_x_2;
15     target_y ← intersection_y_2;
16 new_target_x, new_target_y ← nearest_grid_point(target_x, target_y);
17 connecting_line_segment ← get_line_segment_waypoints(ux, uy, new_target_x,
     new_target_y)[: −1];
18 new_global_path ← create_new_global_path((new_target_x,
     new_target_y), goalNode);
19 USV_index ← USV_path.index((USV_x, USV_y));
20 del USV_path[USV_index : len(USV_path)];
21 USV_path ← USV_path + connecting_line_segment + new_global_path;
22 return USV_path, new_target_x, new_target_y;
```

## 8. Simulation results

The simulations show that this approach can be effective. In **Figures 11** through **15**, avoiding ob1 required two steps to the left, avoiding ob2 and ob4 required two steps to the right, and avoiding ob3 and ob5 required one step to the left. The yellow line in the figures represents an implementation of D* [4]. It can be seen that the local planner effectively detects the direction of the moving obstacle and routes the USV behind it, whereas D* does not.

Thus, these simulation experiments show that this new local path planning algorithm is very simple and stable, and can perform in real-time. It allows the USV to do on-board path replanning as necessary to avoid unexpected obstacles. Also, it is well-integrated with the new global path planner. This is important for a USV performing missions in a complex environment, where the path planner must satisfy both the requirement of real-time on-board local path planning and at the same time meet the demand for complete global path planning.

**Algorithm 12:** $go_left_two_steps$

**input** : $USV_x, USV_y, pux, puy, ux, uy, ox, oy,$ $USV_path$

**output** : $USV_path, target_x, target_y$

1: $slopeOfUSVObLink \leftarrow (oy - uy)/(ox - ux)$;
2: $slopeOfPerpendicular \leftarrow -1/slopeOfUSVObLink$;
3: $offset \leftarrow offsetFactor * min_fixed_distance$;
4: {intersection points of perpendicular with circle around ob};
5: $intersection_x_1 \leftarrow ox + offset/np.sqrt(1 + slopeOfPerpendicular * *2)$;
6: $intersection_y_1 \leftarrow slopeOfPerpendicular * (intersection_x_1 - ox) + oy$;
7: $intersection_x_2 \leftarrow ox - offset/np.sqrt(1 + slopeOfPerpendicular * *2)$;
8: $intersection_y_2 \leftarrow slopeOfPerpendicular * (intersection_x_2 - ox) + oy$;
9: {choose intersection point to left of line from USV to ob};
10: **if** $sign((ux, uy), (ox, oy), (intersection_x_1, intersection_y_1)) > 0$ **then**
11: $target_1_x \leftarrow intersection_x_1$;
12: $target_1_y \leftarrow intersection_y_1$;
13: **else**
14: $target_1_x \leftarrow intersection_x_2$;
15: $target_1_y \leftarrow intersection_y_2$;
16: **end if**
17: $line_segment_1 \leftarrow get_line_segment_waypoints(ux, uy, target_1_x, target_1_y)[: -1]$
18: {intersection points of USV-ob line with circle around ob};
19: $intersection_x_3 \leftarrow ox + offset/np.sqrt(1 + slopeOfUSVObLink * *2)$;
20: $intersection_y_3 \leftarrow slopeOfUSVObLink * (intersection_x_3 - ox) + oy$;
21: $intersection_x_4 \leftarrow ox - offset/np.sqrt(1 + slopeOfUSVObLink * *2)$;
22: $intersection_y_4 \leftarrow slopeOfUSVObLink * (intersection_x_4 - ox) + oy$;
23: {choose intersection point to left of line from target_1 to ob};
24: **if** $sign((target_1_x, target_1_y), (ox, oy), (intersection_x_3, intersection_y_3)) > 0$ **then**
25: $target_2_x \leftarrow intersection_x_3$;
26: $target_2_y \leftarrow intersection_y_3$;
27: **else**
28: $target_2_x \leftarrow intersection_x_4$;
29: $target_2_y \leftarrow intersection_y_4$;
30: **end if**
31: $new_target_2_x, new_target_2_y \leftarrow nearest_grid_point(target_2_x, target_2_y)$;
32: $line_segment_2 \leftarrow get_line_segment_waypoints(target_1_x, target_1_y, new_target_2_x, new_target_2_y)[: -1]$
33: $new_global_path \leftarrow create_new_global_path((new_target_2_x, new_target_2_y), goalNode)$;
34: $USV_index \leftarrow USV_path.index((USV_x, USV_y))$;
35: **del** $USV_path[USV_index : len(USV_path)]$;
36: $USV_path \leftarrow USV_path + line_segment_1 + line_segment_2 + new_global_path$;
37: **return** $new_target_2_x, new_target_2_y$;

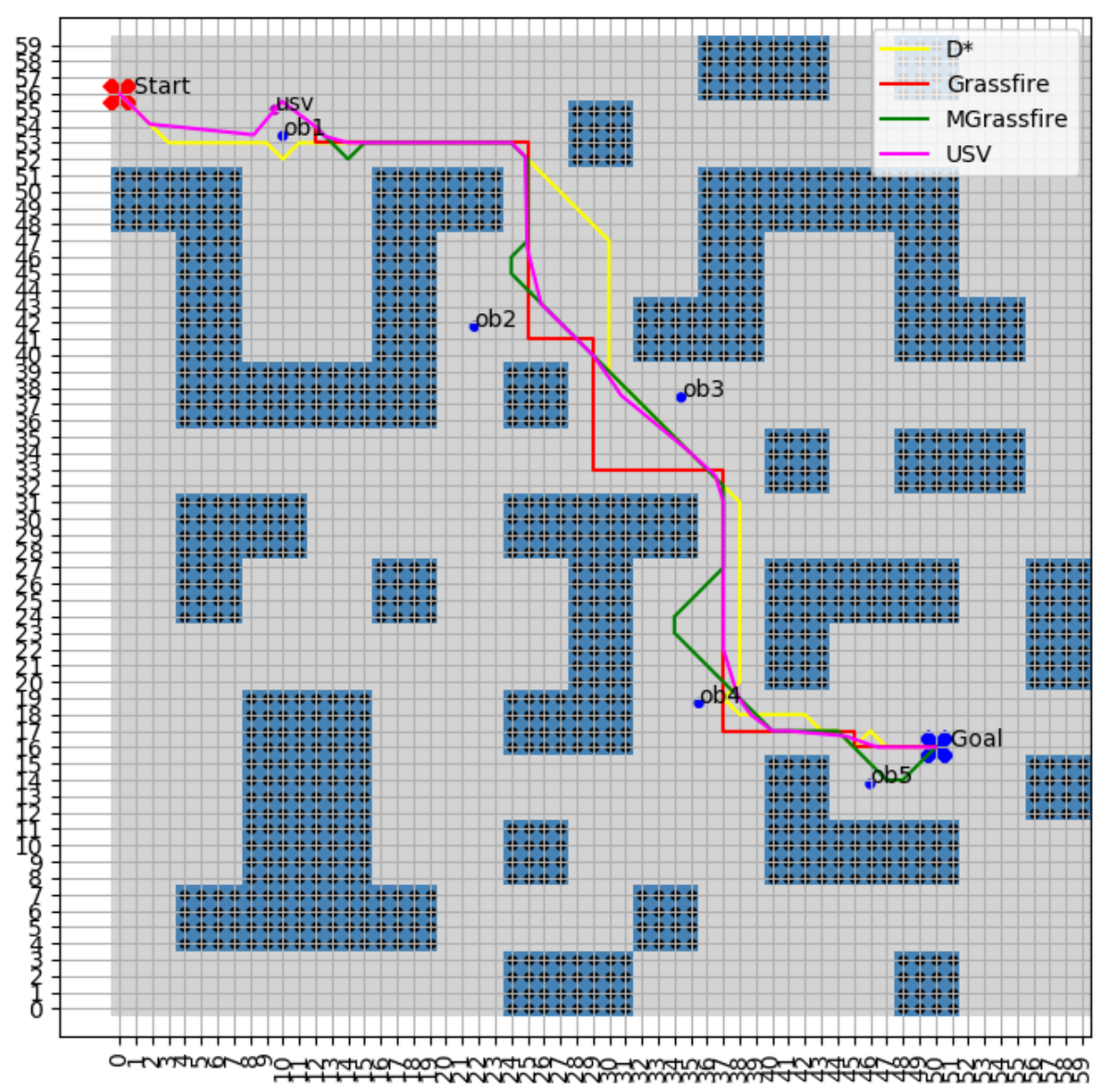

**Figure 11.** USV going to the left behind and around moving obstacle $ob_1$.

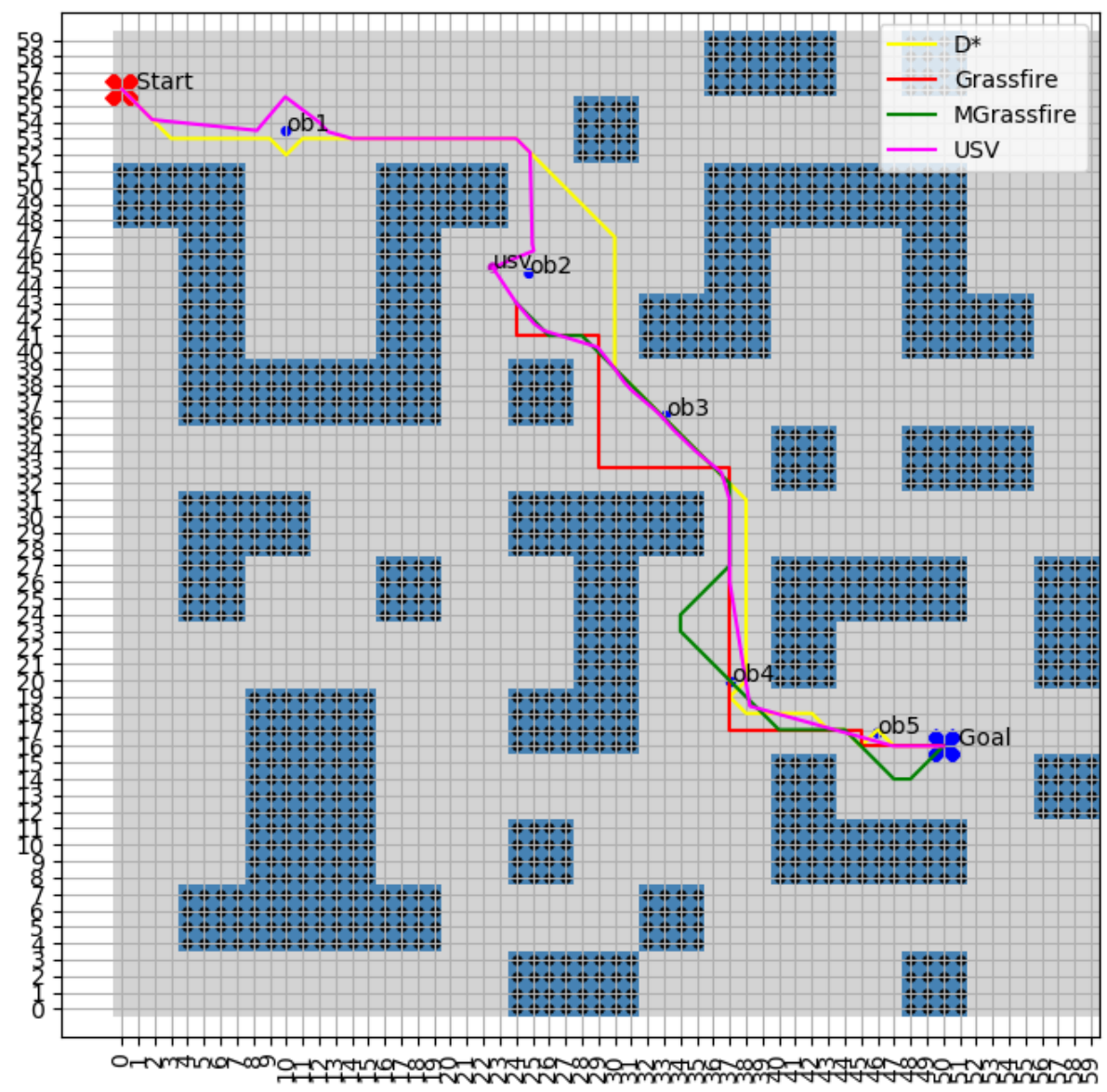

**Figure 12.** USV going to the right behind and around moving obstacle $ob_2$.

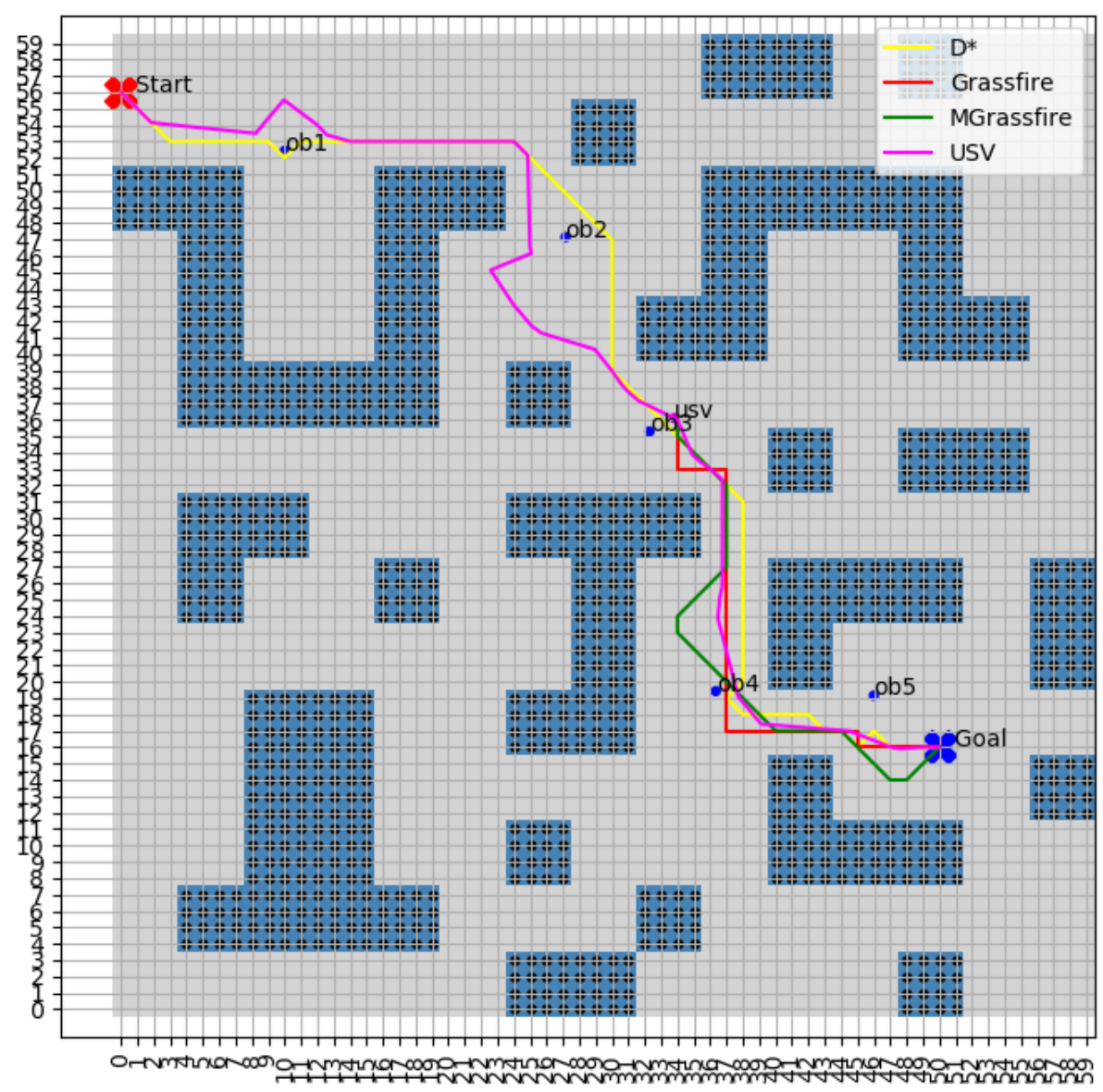

**Figure 13.** USV going to the left behind and around moving obstacle $ob_3$.

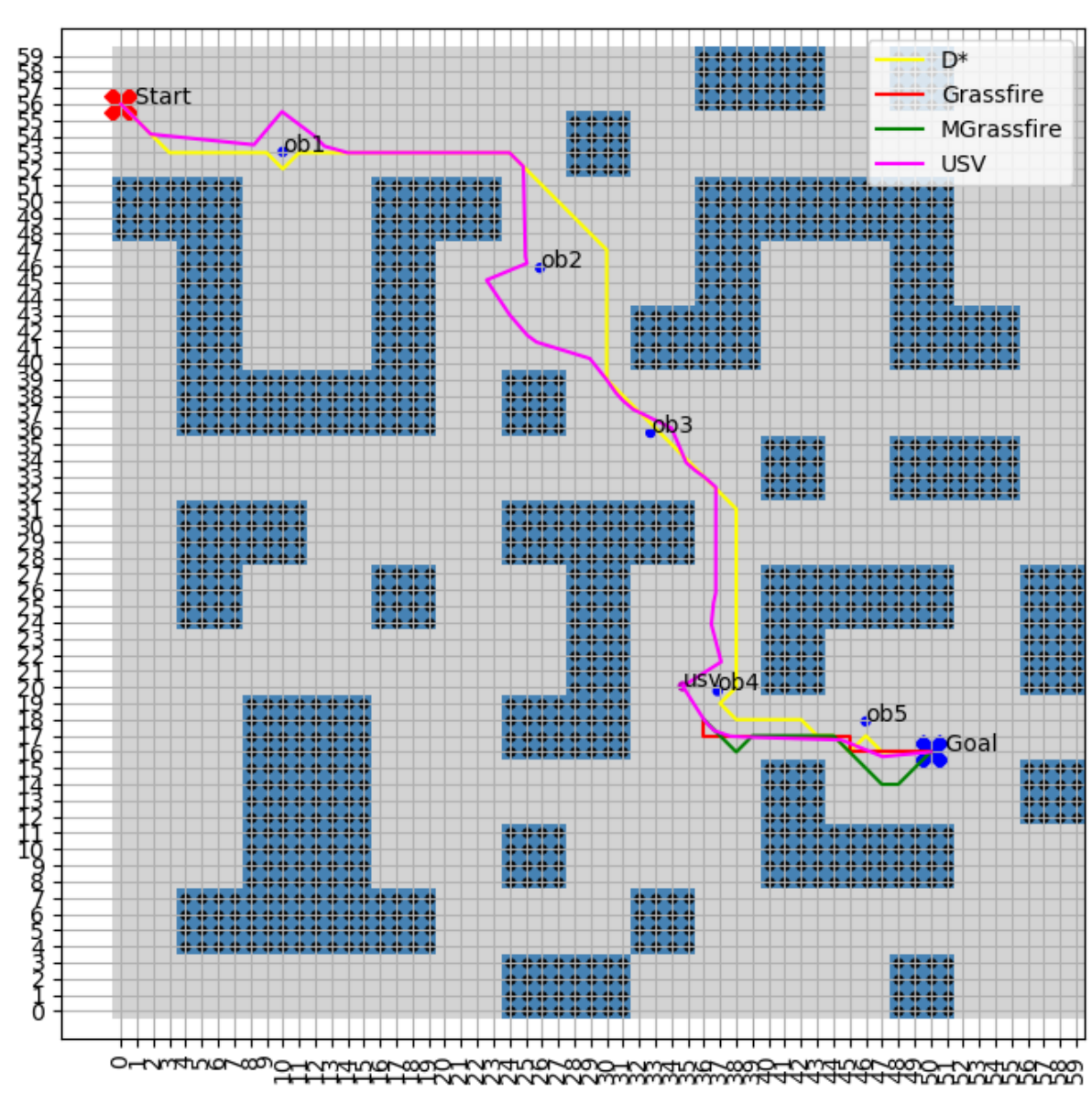

**Figure 14.** USV going to the right behind and around moving obstacle $ob_4$.

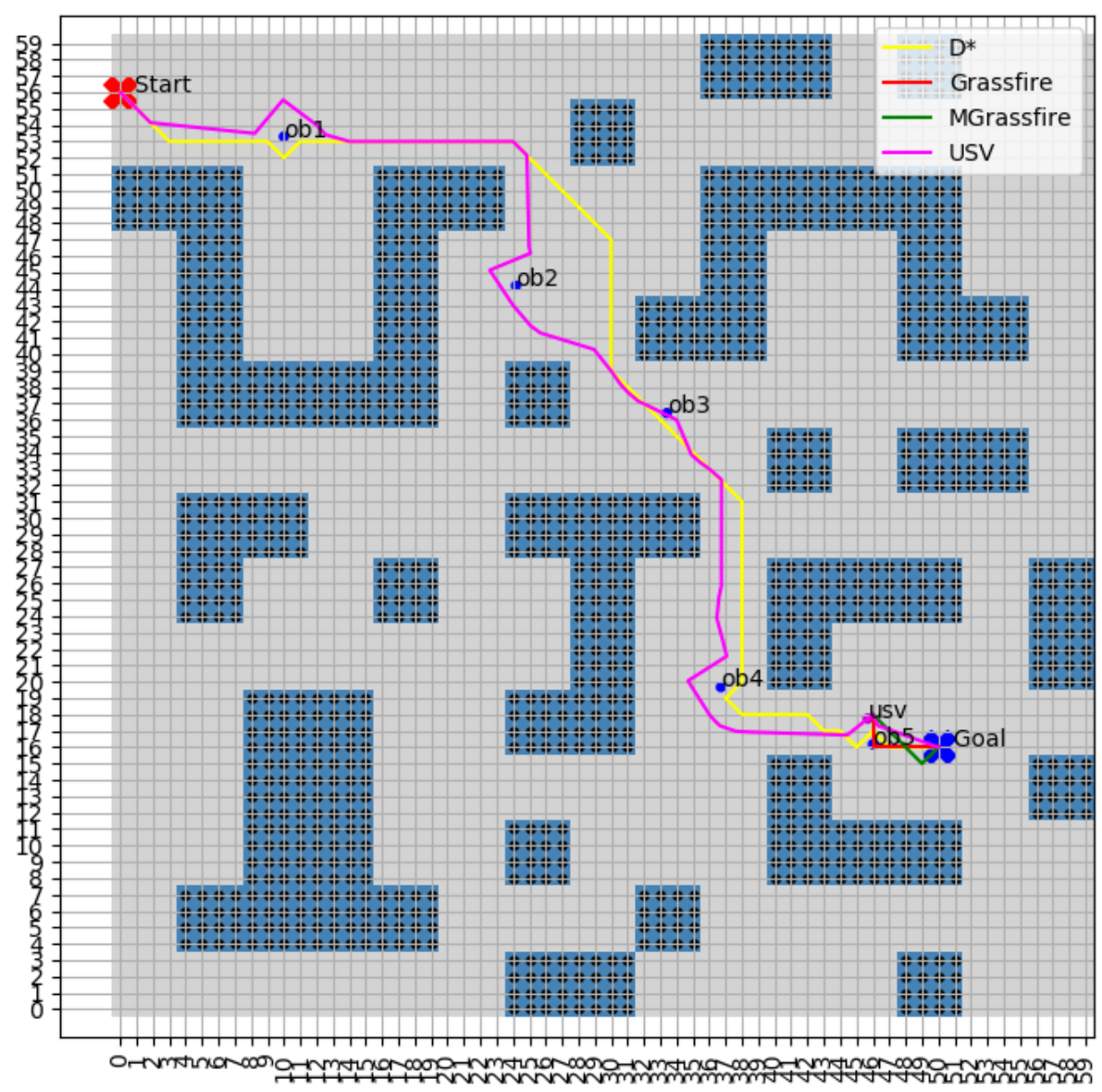

**Figure 15.** USV going to the left behind and around moving obstacle $ob_5$.

## 9. Prospects for future research

There are several ways in which the current work can be extended.

- Generalize to 3D. The present system works with a 2D environment, and thus applies primarily to surface vehicles, but could also apply to underwater vehicles (UUVs) operating at a constant depth. In order to fully accommodate UUVs, however, the model must incorporate a depth measurement. Moreover, in doing so, one should consider that in avoiding an obstacle, an USV can go under the obstacle rather than around it. The decision logic should then include parameters that determine which alternative to take.
- Expand the decision logic. In the Introduction, Section 1, there was mention of the MOOS-IvP system and the associated algorithms presented in [7]. The proposed expansion would include incorporating those algorithms into the local planner. For example, the proper behavior for the USV when it is overtaking a moving obstacle. This, in particular, requires incorporating relative speeds into the decision process.
- Extract the decision logic component into a separate module. In the present simulations, the decision logic is hard coded into the software. This needs to be abstracted into a separate module with an interface that allows the human user to specify the relevant parameters, e.g., the distance between the USV and obstacle at which the circumnavigation process is triggered, or the relative speeds at which the USV decides to go behind the obstacle.
- Avoid multiple moving obstacles. The current system works with one moving obstacle at a time. This needs to be generalized to work with multiple obstacles. It is hereby acknowledged that this is a very complex problem and that there likely is no optimal solution. But it is a problem that somehow needs to be addressed.

## 10. Conclusion

Path planning for USVs in an operational environment having fixed obstacles typically entails computing the path off-line and loading this into the USV before it is launched. This approach is called global planning and requires environment information before the mission execution. This paper began with a presentation of a new global path planning algorithm, one that is guaranteed to find a path from the start node to the goal, as long as such a path is possible, and which produces a path that is smooth (has no sharp turns). This uses a combination of three path planning algorithms, including a new recursive version of the well-known Probabilistic Road Map. It is argued that this version is conceptually simpler, easier to understand, and usually faster. Next the paper presented a local path planner that can detect a moving obstacle and effectively circumnavigate it so as to avoid a collision, including going behind the obstacle whenever this is appropriate. This employs a decision logic based on observations of the relative position and trajectories of the USV and obstacle. The ability to route the USV behind the obstacle distinguishes this local path planner from numerous other path planners, including the well-known D* algorithm. We believe that the use of a higher-level decision logic to accomplish this is new, and that it can be further extended in various ways. In addition, whereas this work has dealt only with a two-dimensional operational environment, and thus only directly applies to surface vehicles, that same methodology as employed here can easily be generalized to three dimensions and thus can apply to both aerial and underwater vehicles.

**Acknowledgments:** This work is the product of a collaboration with my former doctoral student, Mohamad Imran Chowdhury. His current employment does not entail academic research, and he has declined to participate in this publication.

**Conflict of interest:** The author declares no conflict of interest.

## Notes

[1] The original paper is somewhat poorly written, requiring some "reading between the lines" to decipher unexplained terminology and incoherent instructions. In particular, the "if" statement on line (8) of the original pseudocode, p. 569, has a condition that mentions the notion of a "connected component", whereas nowhere in the paper is it explained what this means or what role it plays in creating the road map. In our version of the algorithm, this condition has been removed. Based on the simulation results, I believe that this has implemented the PRM method correctly and as originally intended. It may also be noted that in choosing neighbors in $N$ of a sample point $p$, [2] selects those within a given minimum distance from $p$, but it is mentioned in the paper that an alternative is to choose the $k$ nearest neighbors of $p$ from some $k$. We have taken the latter approach in order to use the KD-tree data structure [9], which simplifies, and increases the efficiency of, the software implementation.